\documentclass{article} %
\PassOptionsToPackage{dvipsnames}{xcolor}
\usepackage{iclr2027_conference,times}

\usepackage{amsmath,amsfonts,bm}

\def\eqref#1{equation~\ref{#1}}

\def\1{\bm{1}}

\DeclareMathAlphabet{\mathsfit}{\encodingdefault}{\sfdefault}{m}{sl}
\SetMathAlphabet{\mathsfit}{bold}{\encodingdefault}{\sfdefault}{bx}{n}

\usepackage{graphicx}
\graphicspath{{figures/images/}}
\usepackage{booktabs} %
\usepackage{wrapfig} %

\usepackage{titletoc} %
 
\definecolor{cvprblue}{rgb}{0.21,0.49,0.74}
\usepackage[pagebackref,breaklinks,colorlinks,citecolor=cvprblue, linkcolor=cvprblue]{hyperref}
\usepackage{cleveref} 
\usepackage{url}
\usepackage{xcolor} %

\newcommand{\parag}[1]{\par\noindent\textbf{#1}\hspace{0.5em}}

\usepackage{xspace}
\usepackage{capt-of} %
\newcommand{\ours}{\texttt{Where-OPD}\xspace}

\title{\ours: Spatially Guided On-Policy Self-Distillation of MLLMs with Synthetic Scenes}

\author{%
Sophia Sirko-Galouchenko$^{1,2}$ \quad
Monika Wysocza\'nska$^{1}$ \quad
Andrei Bursuc$^{1}$ \\
\textbf{Nicolas Thome$^{4,3,2}$ \quad Spyros Gidaris$^{1}$} \\[4pt]
$^{1}$Valeo.ai \quad
$^{2}$Sorbonne Universit\'e, CNRS, ISIR, F-75005 Paris, France \\
$^{3}$Institut universitaire de France (IUF) \quad $^{4}$ ILLS, CNRS, Montreal, QC H2S 3H1
}

\usepackage{colortbl}
\usepackage{siunitx}
\usepackage{pgfplots}
\pgfplotsset{compat=1.17}
\usepgfplotslibrary{groupplots}
\usetikzlibrary{arrows.meta, positioning, calc, fit, backgrounds}
\usepackage{standalone}
\definecolor{deltaup}{RGB}{32,122,78}
\definecolor{deltadn}{RGB}{190,48,40}
\definecolor{oursrow}{RGB}{226,238,251}
\colorlet{baserow}{black!6}
\definecolor{avgcol}{RGB}{237,239,243}
\definecolor{seriesblue}{HTML}{2A78D6}
\newcommand{\dsize}{\fontsize{5}{5.5}\selectfont}
\newcommand{\hd}[1]{{\fontsize{5.6}{6.4}\selectfont #1}}
\newcommand{\du}[1]{{\dsize\color{deltaup}$+$#1}}
\newcommand{\dd}[1]{{\dsize\color{deltadn}$-$#1}}
\newcommand{\de}[1]{{\dsize\color{black!40}$\pm$#1}}
\newcommand{\dc}[2]{{\renewcommand{\arraystretch}{1}\begin{tabular}[t]{@{}c@{}}#1\\[-2.2pt]#2\end{tabular}}}
\newcommand{\bb}[1]{{\renewcommand{\arraystretch}{1}\begin{tabular}[t]{@{}l@{}}#1\\[-2.2pt]{\dsize\color{black!45}base model}\end{tabular}}}

\iclrfinalcopy %
\begin{document}

\maketitle
\lhead{Preprint} %
\vspace{-14pt} %

\begin{abstract}
On-policy self-distillation has recently emerged as an effective approach for improving language-model reasoning by supervising students with a frozen or EMA version of themselves that receives privileged information. Its application to multimodal large language models (MLLMs), however, remains largely unexplored. Recent approaches use privileged visual information, such as image crops corresponding to a question, to improve fine-grained perception, but their gains are 
confined to tasks that benefit from such visual zooming and require either human-annotated grounding data or external teacher models.
We introduce a different form of on-policy self-distillation for MLLMs that provides the teacher with textual, spatially grounded guidance identifying the visual elements relevant to a query. We use procedurally generated scenes with automatically available object identities and spatial coordinates, enabling scalable and annotation-free post-training. The teacher uses this spatial guidance to locate and integrate evidence from multiple relevant image regions, while the student learns to reproduce the resulting behavior from the image and question alone.
Our approach consistently improves performance on counting, document and chart understanding benchmarks across multiple models. 
Importantly, although post-training uses only synthetic scenes, the resulting improvements transfer to real-world perception benchmarks, yielding a 3.23-point gain in average performance across CVBench, V*, ZoomBench, BLINK, HR-Bench, and MME-RealWorld. These results show that spatially grounded privileged information can induce broader perceptual capabilities through on-policy self-distillation, enabling substantial synthetic-to-real transfer beyond the task and data distribution used for post-training.  \href{https://github.com/sirkosophia/Where-OPD}{Project page}

\end{abstract}

\begin{center}
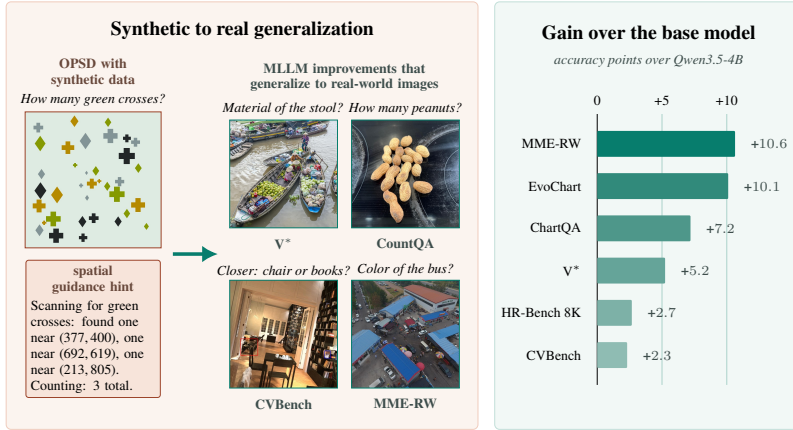

\resizebox{0.76\linewidth}{!}{\includestandalone{figures/teaser_tikz}}
\captionof{figure}{\textbf{Synthetic-to-real transfer of our spatially grounded on-policy self-distillation.}
\textbf{Left:} Our method \ours distills behavior induced by spatially grounded guidance on synthetic counting scenes, yielding improvements on diverse visual perception tasks involving real-world images.
\textbf{Right:} Accuracy gains of \ours over the Qwen3.5-4B base model across six benchmarks, demonstrating transfer beyond the synthetic images and counting task used for post-training.}
\label{fig:teaser}
\end{center}

\section{Introduction}
\label{sec:intro}

On-policy distillation (OPD) \citep{agarwal2024policy} has emerged as an effective approach for post-training large language models. Unlike off-policy distillation, OPD supervises student-generated trajectories with teacher predictions at the states the student actually visits, better aligning training with inference-time behavior. On-policy self-distillation (OPSD) \citep{zhao2026self} extends this paradigm by using a frozen or exponential-moving-average (EMA) copy of the student as a teacher, augmented with privileged information unavailable to the student. The student thus learns from a more informed version of itself while remaining fully self-contained at inference time.

How to exploit this paradigm for multimodal large language models (MLLMs), particularly for improving visual perception, remains largely unexplored. A central question is therefore \emph{what privileged information should be provided to the teacher?} In the multimodal setting, this privileged information can take different forms, ranging from enhanced or localized visual observations to explicit information about the image content. The choice is important because it determines what advantage the teacher has over the student and, consequently, what capabilities can be transferred through self-distillation.

Recent work on OPSD for MLLMs has primarily created this teacher--student difference by giving the teacher better visual access to the image. Vision-OPD~\citep{yuan2026vision} and Imagine-OPD~\citep{cai2026thinking}, for example, provide the teacher with cropped or zoomed-in views of question-relevant regions while the student operates on the original image. Other approaches construct this difference through image resolution or perturbations \citep{zhu2026rp,li2026self}, or contrast teacher predictions under ``positive'' and ``negative'' visual views to derive a more visually grounded distillation signal \citep{aniri2026opd,liang2026visual}. Despite these different implementations, they largely share the same principle: the teacher is given more informative visual observations than the student. Such approaches can yield strong improvements when fine-grained or localized visual evidence is critical, but these gains do not necessarily transfer uniformly across visual tasks. In our evaluation, for instance, Vision-OPD improves Qwen3.5-4B by 8.90 points on V* and 14.56 points on ZoomBench, yet decreases performance on CountQA by 10.73 points. This motivates exploring forms of privileged information that may support broader improvements in visual perception.

This observation raises a different question: \emph{can privileged information specify not what the teacher should see, but where in the image the relevant evidence can be found?} Many visual tasks require identifying question-relevant elements distributed across an image and using them to produce an answer \citep{hudson2019gqa,johnson2017clevr}. Counting requires finding all instances matching a query; reading may require collecting text from different regions; and chart understanding often requires associating labels, values, and graphical elements \citep{masry2022chartqa}. For such tasks, knowing which visual elements are relevant and where they are located could provide useful guidance without giving the teacher a different view of the image. We therefore investigate \emph{spatially grounded privileged information}: textual guidance that identifies question-relevant visual elements and points to their locations in the image.

A practical challenge is obtaining such guidance at scale. Existing multimodal OPSD approaches~\citep{yuan2026vision,cai2026thinking} rely on human-annotated grounding information or external models to identify relevant visual regions. Instead, we construct procedurally generated scenes whose object identities, attributes, and spatial coordinates are known by design, allowing us to automatically generate \emph{spatially grounded hints} that identify question-relevant elements and their locations, without pre-existing datasets, human annotations, or a separate higher-capacity teacher model. During post-training, teacher and student receive the same image and question, but only the teacher receives the hint; the student learns from its behavior and operates without hints at inference time. Our central hypothesis is that \emph{distilling behavior induced by spatially grounded guidance can improve how an MLLM identifies and uses relevant visual evidence, and that these improvements can transfer beyond the synthetic scenes used for post-training.}

We evaluate this approach across three MLLMs---Qwen3.5-4B, Qwen3.5-9B, and Qwen3-VL-4B---and across both task-specific and general visual perception benchmarks. Our method consistently improves performance on counting, reading, and chart-understanding tasks, including CountQA, DocVQA, OCRBench, ChartQA, and EvoChart. For example, on Qwen3.5-4B, it improves ChartQA and EvoChart by 7.20 and 10.11 points, respectively, while also improving CountQA and OCRBench by 2.53 and 1.93 points. More importantly, the benefits are not confined to the synthetic visual distribution used for post-training. Average performance across CVBench, V*, HR-Bench 4K, HR-Bench 8K, ZoomBench, MME-RealWorld, and BLINK improves by 3.23, 1.07, and 1.29 points for the three MLLMs, respectively. \emph{These results indicate that spatially grounded privileged guidance can induce improvements that transfer from simple procedurally generated scenes to substantially different real-world visual tasks.}

Our contributions are threefold:
\begin{itemize}

\item \textbf{Spatially grounded privileged guidance for OPSD.} We introduce a form of on-policy self-distillation in which the teacher receives textual hints identifying question-relevant visual elements and their spatial locations, while the student receives only the image and question.
\item \textbf{Annotation-free generation of spatially grounded hints.} We construct the entire post-training data procedurally, using scene metadata to automatically generate question-relevant spatial guidance without pre-existing datasets, human annotations, or external teacher models.
\item \textbf{Broad improvements and synthetic-to-real transfer.} Across three MLLMs, our approach improves counting, reading, and chart-understanding benchmarks, while also improving general visual perception on real-world images, demonstrating transfer beyond both the tasks and visual distribution used for post-training.
\end{itemize}
\section{Related Works}

\paragraph{Improving visual reasoning in MLLMs.} Efforts to strengthen visual understanding and reasoning in MLLMs can be broadly grouped by the training stage they target. At the pretraining stage, a large body of work focuses on modifying the architecture to better expose visual information to the language model \citep{mckinzie2024mm1, liu2024improved, cha2024honeybee, chen2024lion, lin2025multi, kar2024brave, tong2024cambrian, azadani2025leo, shi2024eagle, lu2025deepseek}. Other works trace the bottleneck to how the LLM uses visual information during decoding \citep{fu2025hidden}, and introduces auxiliary objectives to counteract it \citep{wang2025reconstructive, yoon2025visual, caffagni2025seeing}.
Another line of work instead improves visual capabilities through instruction-tuning data. V-GIFT \citep{sirko2026boosting} shows that augmenting the instruction data mix with  visually grounded, self-supervised-style tasks is sufficient to improve visual perception. Related efforts construct vision-centric instruction data through dense, detailed captions \citep{chen2024sharegpt4v}, region-level and grounded conversations \citep{chen2023shikra}, or synthetic data targeting fine-grained visual differences \citep{jiao2025img}.
More recently visual capabilities can be improved during post-training. For example GRPO-style methods have been successfully adapted to multimodal models \citep{huang2026vision, Liu_2025_ICCV, yu2026perception}, including approaches that use self-supervised visual tasks such as jigsaw puzzles as verifiable training signals \citep{wang2025jigsawr1}. 
However, such methods provide only sparse, sequence-level rewards, which offer limited guidance on which parts of a response fail to use the visual input. This has motivated a shift toward denser, token-level supervision through on-policy self-distillation (OPSD), which we discuss next.

\paragraph{Visual On-policy Self-Distillation.} On-policy distillation (OPD) has emerged as an effective technique for improving reasoning in language models and, more recently, in multimodal models. It relies on token-level supervision from a stronger teacher \citep{agarwal2024policy}, while on-policy self-distillation removes the need for an external teacher by deriving the teacher signal from the model itself under privileged context \citep{zhao2026self}. A recent surge of concurrent approaches for MLLMs differs mainly in the type of the teacher–student asymmetry. For example Vision-OPD~\citep{yuan2026vision} conditions the teacher on an evidence-centered crop while the student sees a full image. Imagine-OPD~\citep{cai2026thinking} gives the teacher privileged zoomed evidence views and distills them into imagination-based reasoning. OPD-V~\citep{aniri2026opd} uses a positive teacher conditioned on an evidence-centered crop and a negative teacher conditioned on masked crop while RP-OPSD~\citep{zhu2026rp} constructs the teacher-student asymmetry through resolution. 
S$^2$VOPD~\citep{li2026self} takes 
a different approach by removing information from the student rather than adding privileged information to the teacher. The teacher observes the original image while the student observes a strongly augmented version of the same image. VCSD~\citep{liang2026visual} contrasts teacher predictions under the original image and a content-erased control. 
Finally, ViCuR~\citep{tian2026vicur} replaces reasoning-trace-based privilege with question-relevant visual cues that describe evidence already present in the image. In this work, we also induce the teacher--student asymmetry by providing the teacher with visually grounded hints expressed in text. Inspired by prior work on procedurally generated data for visual and spatial reasoning with programmatically generated questions \citep{johnson2017clevr, hudson2019gqa} we construct the entire post-training dataset procedurally, using scene metadata to automatically generate question-relevant spatial guidance.

\section{Method}
\label{sec:method}

\begin{figure}[t]
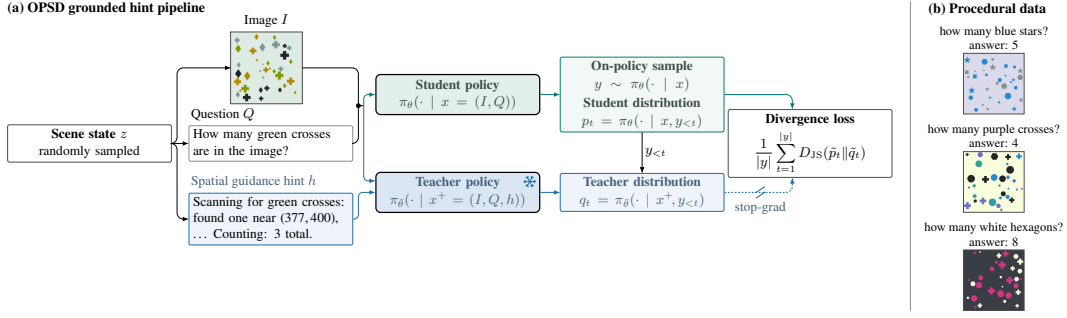

\centering
\resizebox{\linewidth}{!}{\includestandalone{figures/method_tikz}}
\caption{\textbf{Spatially grounded privileged guidance with \ours{}.}
\textbf{(a)} A procedural scene provides an image $I$, a question $Q$,
and a hint $h$ listing the target objects' locations. The student receives
$x=(I,Q)$; the frozen teacher receives $x^{+}=(I,Q,h)$ and scores the
student's sampled prefixes for on-policy distillation. Only the student's parameters are
updated. \textbf{(b)} Visualization of training scenes showing variation in
backgrounds, object categories, and counts.}
\label{fig:method}
\end{figure}

We introduce an on-policy self-distillation framework in which the teacher receives textual, spatially grounded guidance identifying question-relevant visual elements and their locations, while the student observes only the image
and question (see \autoref{fig:method} for an overview). We obtain this privileged information automatically from procedurally generated scenes, enabling annotation-free post-training without an external teacher model. We first
review on-policy self-distillation (\autoref{sec:prelim}), then present our spatially grounded guidance, its procedural generation, and the resulting training objective (\autoref{sec:guidance}).

\subsection{Preliminaries: On-Policy Self-Distillation}
\label{sec:prelim}

On-policy self-distillation (OPSD) uses a frozen or exponential-moving-average (EMA) copy of the student as a teacher, augmented with privileged information unavailable to the student. 
Let $\pi_\theta$ denote the student model and $\pi_{\bar{\theta}}$ the privileged teacher model. Given an input $x$, which for an MLLM consists of an image $I$ and a question $Q$, i.e., $x=(I,Q)$, on-policy distillation (OPD) samples trajectories from the current student policy,
\begin{equation}
    y \sim \pi_\theta(\cdot \mid x).
\end{equation}
The teacher then provides token-level supervision on the same prefixes visited by the student. At token position $t$, their predictive distributions are
\begin{equation}
    p_t = \pi_\theta(\cdot \mid x, y_{<t}),
    \qquad
    q_t = \pi_{\bar{\theta}}(\cdot \mid x^{+}, y_{<t}).
\end{equation}
where $x^{+}$ denotes the privileged input available only to the teacher. Distillation is thus performed by minimizing a divergence between $p_t$ and $q_t$ on states visited by the current student policy (on-policy), rather than on trajectories generated independently by the teacher.

For MLLMs, the construction of $x^{+}$ determines the teacher--student information gap and is therefore a central design choice. Prior work commonly provides the teacher with an enhanced or localized visual observation $I^{+}$, i.e., $x^{+}=(I^{t},Q)$, while the student observes the original image $I$. We instead keep the visual observation unchanged and provide the teacher with textual, spatially grounded privileged information, $h$ as described next.

\subsection{Spatially Grounded Privileged Guidance}
\label{sec:guidance}

Our goal is to provide the teacher with explicit information about \emph{where} the visual evidence relevant to a question is located, rather than with a different view of the image. 
Specifically, given an image $I$, question $Q$, and spatially grounded textual hint $h$, the student receives $x=(I,Q)$, while the teacher additionally receives the hint, $x^{+}=(I,Q,h)$.

The hint identifies the visual elements relevant to the question and points to their locations in the image. For example, for a question asking how many cars are present, such guidance could identify the image locations corresponding to the relevant cars and provide their resulting count.  
The teacher therefore has explicit access to the locations of the evidence needed to answer the question, while the student must infer the relevant visual evidence from the image and question alone. The privileged guidance is used only during post-training and is absent at inference time.

\subsubsection{Procedural Generation of Spatial Guidance}
\label{sec:data}

A practical challenge is obtaining spatial guidance at scale without human annotations or an external grounding model. 
We address this by procedurally generating post-training data, such that the identities, attributes, and locations of all visual elements are known from the scene-generation process.
We use counting questions because answering them requires identifying all instances matching a query, naturally providing supervision over multiple question-relevant image locations.

Each training example is built from a \emph{scene}: a synthetic canvas on which
simple visual elements, colored geometric shapes, are placed at
non-overlapping positions over a uniform background.
Unlike a natural image, a scene is fully specified by the generator before it
is rendered.
Formally, a scene is described by its state
\begin{equation}
z = \bigl(\{(c_i,r_i,u_i,v_i)\}_{i=1}^{M}, b\bigr),
\end{equation}
where $c_i$ denotes the color--shape category of the $i$-th visual element,
$r_i$ its radius, $(u_i,v_i)$ its center in image coordinates, $M$ the number
of visual elements, and $b$ the background color.
The state thus records the identity, attributes, and location of every
element, and rendering $z$ produces the corresponding image $I$.

For each scene, we sample a counting question $Q$ targeting a category $c$ present in $z$. Since the complete scene specification is known, we can directly identify all question-relevant elements as
\begin{equation}
\mathcal{O}_c(z) = \{(u_i,v_i) \mid c_i = c\},
\end{equation}
whose count $N_c = |\mathcal{O}_c(z)|$ gives the answer $a$. We then construct the privileged textual hint using a fixed template $\mathcal{T}$,
\begin{equation}
h = \mathcal{T}\bigl(c,\mathcal{O}_c(z)\bigr),
\label{eq:hint}
\end{equation}
which names the target category, lists the location of each matching instance, and concludes with their count. 
For example, for the question \emph{``How many green crosses are there?''}, if two matching objects occur at $(377,400)$ and $(213,805)$, the privileged hint is:
\begin{quote}\small\ttfamily
 Scanning for green crosses: found one near (377,\,400),
  found one near (213,\,805). Counting: 2 total.
\end{quote}

Thus, while the post-training task itself is simple, its privileged signal explicitly identifies multiple pieces of question-relevant visual evidence and their spatial locations. 
Notably, the complete training tuple $(I,Q,a,h)$ is generated automatically from $z$, without human annotations, region proposals, or an external higher-capacity MLLM.

We generate $1024\times1024$ scenes containing colored geometric objects from seven shape classes and twelve colors on a flat background. Each scene contains between 12 and 40 objects, each drawn with a radius between 14 and 64 pixels. 
Questions ask for the number of objects belonging to a sampled color--shape category. Our main post-training set contains 3,000 generated image--question pairs with their corresponding privileged hints. More details about the generation can be found in~\ref{sec:supp:synthdata} of the Appendix.

\subsubsection{Spatially Privileged On-Policy Self-Distillation}
\label{sec:opsd}

We use the procedurally generated hints within OPSD. Both student and teacher are initialized from the same pretrained checkpoint. In our main setting, the teacher is frozen at initialization, $\bar{\theta}=\theta_0$.
For each training example, we instantiate the student and privileged teacher inputs as $x=(I,Q)$ and $x^{+}=(I,Q,h)$, respectively, and apply on-policy self-distillation as described in \autoref{sec:prelim}.

We minimize the generalized Jensen--Shannon divergence between the student and privileged teacher distributions, with $\beta=0.5$.
For computational efficiency, the divergence is computed over the
student's top $K=100$ tokens and the corresponding teacher logits, alongside a
tail-probability term.
Let $\tilde p_t$ and $\tilde q_t$ denote the resulting distributions. Our training objective is
\begin{equation}
    \mathcal{L}_{\mathrm{distill}}
    =
    \mathbb{E}_{y \sim \pi_\theta(\cdot \mid x)}
    \Biggl[
    \frac{1}{|y|}
    \sum_{t=1}^{|y|}
    D_{\mathrm{JS}}
    \bigl(\tilde p_t \,\Vert\, \tilde q_t\bigr)
    \Biggr],
    \label{eq:distill}
\end{equation}
where
\begin{equation}
    D_{\mathrm{JS}}(p \Vert q)
    =
    \frac{1}{2}D_{\mathrm{KL}}(p \Vert m)
    +
    \frac{1}{2}D_{\mathrm{KL}}(q \Vert m),
    \qquad
    m = \frac{1}{2}(p+q).
\end{equation}
The privileged teacher distribution $\tilde q_t$ is treated as a fixed target, so gradients flow only through the student. At inference time, the model receives only $(I,Q)$; the privileged hint is used exclusively during post-training.
\section{Experiments}
\label{sec:experiments}

In this section, we evaluate whether spatially grounded privileged guidance improves visual capabilities of MLLMs.
We first describe the experimental setup in~\autoref{sec:exp:setup} and compare with prior on-policy self-distillation methods in~\autoref{sec:exp:main}. Finally, we analyze the training signal and our design choices in~\autoref{sec:exp:analysis}.

\subsection{Experimental Setup}
\label{sec:exp:setup}

\parag{Training details.}
Our experimental setup covers three base models: Qwen3.5-4B and Qwen3.5-9B~\citep{team2026qwen3}, as well as Qwen3-VL-4B~\citep{bai2025qwen3}. We initialize a teacher and a student from the same checkpoint and train on $3{,}000$ procedurally generated counting scenes. For smaller models (Qwen3.5-4B and Qwen3-VL-4B) we apply \ours~ with a multiple-choice question protocol, while for Qwen3.5-9B we implement an open-ended version for same scenes. Detailed input can be found in~\autoref{fig:train_input}. All models are trained for one epoch. Only \ours results for  Qwen3.5-4B and Qwen3.5-9B in~\autoref{tab:main_results} are each averaged over three independent training runs with different fixed seeds; all other post-training results, including the ablations and analyses, come from one run per setting. Further details are given in ~\autoref{sup:training} of the Appendix.

\parag{Baselines.}
We compare \ours{} with Vision-OPD~\citep{yuan2026vision}, OPD-V~\citep{aniri2026opd}, Imagine-OPD~\citep{cai2026thinking}, and S$^2$VOPD~\citep{li2026self}. We use publicly released or author-provided checkpoints and evaluate them with the same protocol as our models. 

\parag{Evaluation benchmarks.}
We evaluate on a broad suite of 15 benchmarks spanning a wide range of general visual skills. These include counting with CountQA~\citep{tamarapalli2025countqa}, document understanding with DocVQA~\citep{mathew2021docvqa} and OCRBench~\citep{liu2024ocrbench}, and chart understanding with ChartQA~\citep{masry2022chartqa} and EvoChart~\citep{huang2025evochart}, as well as general visual perception with CVBench~\citep{tong2024cambrian}, BLINK~\citep{fu2024blink}, MME-RealWorld~\citep{zhang2025mme}, GQA~\citep{hudson2019gqa}, as well as HallusionBench~\citep{guan2024hallusionbench} and AMBER~\citep{wang2023amber} to assess hallucinations. We further evaluate on benchmarks that require fine-grained, detail-level understanding, where answering depends on locating and zooming into small regions of high-resolution images including: V$^*$~\citep{wu2024v}, HR-Bench 4K and 8K~\citep{wang2025divide}, and ZoomBench~\citep{wei2026zooming}. Since the suite consists largely of real-world images and tasks beyond counting, it measures how well skills learned on our rendered training scenes transfer to realistic settings.
Compared with the narrower evaluation suites used in prior work, this broader selection lets us assess generalization across diverse visual capabilities.
We report accuracy (\%) on each benchmark and the unweighted mean across all reported benchmarks. Unless otherwise specified, our evaluation uses greedy decoding with thinking disabled.

\subsection{Main Results}
\label{sec:exp:main}

\begin{table}[t]

\caption{\textbf{Main results across three MLLMs.} Accuracy (\%) on 15 benchmarks; Avg. is their unweighted mean. The smaller
numbers below each score show the change in accuracy relative to the corresponding base model. CVB: CVBench; MME-RW: MME-RealWorld; OCRB: OCRBench; HallB: HallusionBench; HR-4K/8K: HR-Bench 4K/8K;
Zoom: ZoomBench. For \ours on Qwen3.5-4B and Qwen3.5-9B as base model, each score is the mean of three training runs with different seeds; the standard deviation of Avg.\ across runs is $\pm$0.20 for both  Qwen3.5-4B and Qwen3.5-9B.}
\label{tab:main_results}
\centering
\scriptsize
\setlength{\tabcolsep}{1.1pt}
\renewcommand{\arraystretch}{1.15}
\begin{tabular}{l ccccccccccc !{\hspace{5pt}} cccc !{\hspace{5pt}} >{\columncolor{avgcol}}c}
\toprule
 & \multicolumn{11}{c}{General} & \multicolumn{4}{c}{High-resolution} & \multicolumn{1}{c}{} \\
\cmidrule(lr){2-12} \cmidrule(lr){13-16}
Model & \hd{CVB} & \hd{MME-RW} & \hd{BLINK} & \hd{GQA} & \hd{CountQA} & \hd{ChartQA} & \hd{DocVQA} & \hd{OCRB} & \hd{EvoChart} & \hd{HallB} & \hd{AMBER} & \hd{V$^*$} & \hd{HR-4K} & \hd{HR-8K} & \hd{Zoom} & \hd{Avg.} \\
\midrule
\rowcolor{baserow} \bb{Qwen3.5-4B} & 87.45 & 64.11 & 66.28 & 69.32 & 32.00 & 79.32 & 96.00 & 87.30 & 68.32 & 69.40 & 88.80 & 82.72 & \textbf{86.25} & 80.00 & 50.65 & 73.86 \\[2pt]
Vision-OPD & \dc{87.53}{\du{0.08}} & \dc{75.26}{\du{11.15}} & \dc{62.02}{\dd{4.26}} & \dc{68.23}{\dd{1.09}} & \dc{21.27}{\dd{10.73}} & \dc{78.52}{\dd{0.80}} & \dc{95.98}{\dd{0.02}} & \dc{87.00}{\dd{0.30}} & \dc{67.28}{\dd{1.04}} & \dc{65.83}{\dd{3.57}} & \dc{86.32}{\dd{2.48}} & \dc{91.62}{\du{8.90}} & \dc{83.50}{\dd{2.75}} & \dc{80.25}{\du{0.25}} & \dc{65.21}{\du{14.56}} & \dc{74.39}{\du{0.53}} \\[2pt]
OPD-V & \dc{85.97}{\dd{1.48}} & \dc{\textbf{75.76}}{\du{11.65}} & \dc{61.91}{\dd{4.37}} & \dc{67.61}{\dd{1.71}} & \dc{26.11}{\dd{5.89}} & \dc{81.12}{\du{1.80}} & \dc{96.00}{\de{0.00}} & \dc{87.50}{\du{0.20}} & \dc{70.32}{\du{2.00}} & \dc{68.03}{\dd{1.37}} & \dc{82.78}{\dd{6.02}} & \dc{\textbf{94.24}}{\du{11.52}} & \dc{85.00}{\dd{1.25}} & \dc{82.38}{\du{2.38}} & \dc{\textbf{65.44}}{\du{14.79}} & \dc{75.34}{\du{1.48}} \\[2pt]
S$^2$VOPD & \dc{87.91}{\du{0.46}} & \dc{75.59}{\du{11.48}} & \dc{64.70}{\dd{1.58}} & \dc{68.61}{\dd{0.71}} & \dc{31.87}{\dd{0.13}} & \dc{79.88}{\du{0.56}} & \dc{95.98}{\dd{0.02}} & \dc{87.50}{\du{0.20}} & \dc{69.44}{\du{1.12}} & \dc{65.72}{\dd{3.68}} & \dc{88.20}{\dd{0.60}} & \dc{88.48}{\du{5.76}} & \dc{85.12}{\dd{1.13}} & \dc{\textbf{83.88}}{\du{3.88}} & \dc{56.09}{\du{5.44}} & \dc{75.26}{\du{1.40}} \\[2pt]
\rowcolor{oursrow}  \ours & \dc{\textbf{89.76}}{\du{2.31}} & \dc{74.75}{\du{10.64}} & \dc{\textbf{67.77}}{\du{1.49}} & \dc{\textbf{75.41}}{\du{6.09}} & \dc{\textbf{34.53}}{\du{2.53}} & \dc{\textbf{86.52}}{\du{7.20}} & \dc{\textbf{97.00}}{\du{1.00}} & \dc{\textbf{89.23}}{\du{1.93}} & \dc{\textbf{78.43}}{\du{10.11}} & \dc{\textbf{80.16}}{\du{10.76}} & \dc{\textbf{93.28}}{\du{4.48}} & \dc{87.96}{\du{5.24}} & \dc{85.62}{\dd{0.63}} & \dc{82.67}{\du{2.67}} & \dc{51.52}{\du{0.87}} & \dc{\textbf{78.31}}{\du{4.45}} \\[2pt]
\midrule
\rowcolor{baserow} \bb{Qwen3.5-9B} & \textbf{88.02} & 71.45 & \textbf{69.86} & 69.45 & 37.04 & 80.12 & 96.47 & 90.00 & 69.04 & 71.82 & 88.79 & 85.86 & 85.62 & 81.25 & 52.31 & 75.81 \\[2pt]
Vision-OPD & \dc{87.49}{\dd{0.53}} & \dc{73.29}{\du{1.84}} & \dc{66.39}{\dd{3.47}} & \dc{69.27}{\dd{0.18}} & \dc{35.01}{\dd{2.03}} & \dc{81.12}{\du{1.00}} & \dc{\textbf{96.80}}{\du{0.33}} & \dc{89.90}{\dd{0.10}} & \dc{71.44}{\du{2.40}} & \dc{72.98}{\du{1.16}} & \dc{87.74}{\dd{1.05}} & \dc{91.10}{\du{5.24}} & \dc{\textbf{87.38}}{\du{1.76}} & \dc{\textbf{86.75}}{\du{5.50}} & \dc{\textbf{64.97}}{\du{12.66}} & \dc{77.44}{\du{1.64}} \\[2pt]
OPD-V & \dc{84.99}{\dd{3.03}} & \dc{71.58}{\du{0.13}} & \dc{61.13}{\dd{8.73}} & \dc{69.18}{\dd{0.27}} & \dc{30.30}{\dd{6.74}} & \dc{81.16}{\du{1.04}} & \dc{96.67}{\du{0.20}} & \dc{\textbf{90.60}}{\du{0.60}} & \dc{72.64}{\du{3.60}} & \dc{72.34}{\du{0.52}} & \dc{87.68}{\dd{1.11}} & \dc{\textbf{93.72}}{\du{7.86}} & \dc{83.75}{\dd{1.87}} & \dc{83.25}{\du{2.00}} & \dc{63.91}{\du{11.60}} & \dc{76.19}{\du{0.39}} \\[2pt]
\rowcolor{oursrow}  \ours & \dc{87.97}{\dd{0.05}} & \dc{\textbf{74.73}}{\du{3.28}} & \dc{69.24}{\dd{0.62}} & \dc{\textbf{69.52}}{\du{0.07}} & \dc{\textbf{38.26}}{\du{1.22}} & \dc{\textbf{86.05}}{\du{5.93}} & \dc{96.71}{\du{0.24}} & \dc{90.27}{\du{0.27}} & \dc{\textbf{75.47}}{\du{6.43}} & \dc{\textbf{75.39}}{\du{3.57}} & \dc{\textbf{89.06}}{\du{0.27}} & \dc{87.78}{\du{1.92}} & \dc{86.38}{\du{0.76}} & \dc{84.67}{\du{3.42}} & \dc{51.99}{\dd{0.32}} & \dc{\textbf{77.57}}{\du{1.76}} \\[2pt]
\midrule
\rowcolor{baserow} \bb{Qwen3-VL-4B} & 85.44 & 63.22 & 65.49 & 71.01 & 27.68 & 77.20 & \textbf{95.29} & \textbf{87.70} & 62.48 & 70.66 & 89.61 & 81.15 & 80.00 & 75.12 & 45.09 & 71.81 \\[2pt]
OPD-V & \dc{83.89}{\dd{1.55}} & \dc{\textbf{68.10}}{\du{4.88}} & \dc{62.70}{\dd{2.79}} & \dc{70.42}{\dd{0.59}} & \dc{24.74}{\dd{2.94}} & \dc{77.24}{\du{0.04}} & \dc{95.25}{\dd{0.04}} & \dc{86.80}{\dd{0.90}} & \dc{62.48}{\de{0.00}} & \dc{70.14}{\dd{0.52}} & \dc{88.24}{\dd{1.37}} & \dc{\textbf{86.91}}{\du{5.76}} & \dc{81.25}{\du{1.25}} & \dc{\textbf{77.00}}{\du{1.88}} & \dc{\textbf{54.67}}{\du{9.58}} & \dc{72.66}{\du{0.85}} \\[2pt]
Imagine-OPD & \dc{85.97}{\du{0.53}} & \dc{66.38}{\du{3.16}} & \dc{\textbf{66.07}}{\du{0.58}} & \dc{\textbf{71.28}}{\du{0.27}} & \dc{27.29}{\dd{0.39}} & \dc{77.28}{\du{0.08}} & \dc{95.27}{\dd{0.02}} & \dc{87.30}{\dd{0.40}} & \dc{63.76}{\du{1.28}} & \dc{68.35}{\dd{2.31}} & \dc{89.79}{\du{0.18}} & \dc{85.34}{\du{4.19}} & \dc{\textbf{82.25}}{\du{2.25}} & \dc{76.62}{\du{1.50}} & \dc{45.56}{\du{0.47}} & \dc{72.57}{\du{0.76}} \\[2pt]
\rowcolor{oursrow} \ours & \dc{\textbf{86.58}}{\du{1.14}} & \dc{65.25}{\du{2.03}} & \dc{65.86}{\du{0.37}} & \dc{71.16}{\du{0.15}} & \dc{\textbf{30.37}}{\du{2.69}} & \dc{\textbf{77.68}}{\du{0.48}} & \dc{95.27}{\dd{0.02}} & \dc{\textbf{87.70}}{\de{0.00}} & \dc{\textbf{64.24}}{\du{1.76}} & \dc{\textbf{71.29}}{\du{0.63}} & \dc{\textbf{90.06}}{\du{0.45}} & \dc{84.29}{\du{3.14}} & \dc{81.38}{\du{1.38}} & \dc{75.25}{\du{0.13}} & \dc{45.92}{\du{0.83}} & \dc{\textbf{72.82}}{\du{1.01}} \\[2pt]
\bottomrule
\end{tabular}
\vspace{-10pt}
\end{table}

\parag{Broad gains across MLLMs and benchmarks.}
As shown in  \autoref{tab:main_results}, our method achieves the highest average across all evaluated benchmarks on each of the MLLMs. Improvements span multiple visual capabilities: CountQA, ChartQA, EvoChart, and HallusionBench improve consistently across all three MLLMs, while document understanding performance is largely preserved or improved. In contrast, methods based on enhanced or localized visual observations achieve substantial gains on high-resolution visual-search benchmarks, particularly V* and ZoomBench, but show less consistent improvements across other tasks. For instance, Vision-OPD and OPD-V decrease CountQA and BLINK performance on both Qwen3.5-4B and Qwen3.5-9B and often degrade hallucination-sensitive performance. Overall, these results indicate that our spatially grounded guidance yields improvements across a broader range of the evaluated tasks.

\parag{Transfer beyond synthetic counting scenes.}
Despite using only procedurally generated geometric scenes for post-training, our method improves performance on diverse benchmarks involving real-world images and tasks beyond counting. Notably, MME-RealWorld, HallusionBench, $V^{*}$, HR-Bench 8K, GQA, and AMBER improve across all three MLLMs, alongside gains in chart and document understanding. This demonstrates that the benefits of our training approach extend beyond both the visual appearance and the counting task of the synthetic training data.

\subsection{Experimental Analysis}
\label{sec:exp:analysis}
In this section, we present additional experiments that provide further insight into our method. Unless stated otherwise, we report the average accuracy over seven representative benchmarks covering each skill category.

\begin{table}[t]
\caption{\textbf{Training signals and privileged hints on Qwen3.5-4B.}
SFT, GRPO, and the OPSD variants are trained on the same $3,000$ synthetic
scenes for the same number of steps as \ours. Results are accuracy (\%); Avg. is
the unweighted mean across benchmarks present in the table. CVB: CVBench; HR-4K/8K:
HR-Bench 4K/8K; Zoom: ZoomBench.}
\label{tab:training_signal}

\centering
\scriptsize
\setlength{\tabcolsep}{2pt}
\renewcommand{\arraystretch}{1.15}
\begin{tabular}{l ccccccc !{\hspace{5pt}} >{\columncolor{avgcol}}c}
\toprule
Training signal & CVB & V$^*$ & HR-4K & HR-8K & Zoom & CountQA & EvoChart & Avg. \\
\midrule
\rowcolor{baserow} Qwen3.5-4B & 87.45 & 82.72 & 86.25 & 80.00 & 50.65 & 32.00 & 68.32 & 69.63 \\
SFT & 87.68 & \textbf{89.01} & 83.38 & 79.12 & 50.18 & 31.81 & 66.72 & 69.70 \\
GRPO & 87.83 & 83.25 & 81.88 & 80.12 & 49.11 & 33.12 & 71.60 & 69.56 \\
\midrule
OPSD, no privileged information & 87.53 & 82.72 & 85.00 & 81.00 & 50.41 & 32.46 & 68.32 & 69.63 \\
OPSD, cropped-image hint & 88.44 & 83.77 & 82.50 & 79.25 & \textbf{52.31} & 20.22 & 74.80 & 68.76 \\
OPSD, answer-only hint & 88.48 & 84.82 & 83.25 & 79.38 & 51.12 & 33.51 & 78.32 & 71.27 \\
OPSD, spatial guidance hint, no total & 89.88 & 87.96 & 85.25 & 81.00 & 50.77 & \textbf{34.62} & 76.96 & 72.35 \\

\rowcolor{oursrow} OPSD, spatial guidance hint (\ours) & \textbf{89.92} & 88.48 & 85.50 & \textbf{83.25} & 51.01 & 33.77 & \textbf{78.64} & \textbf{72.94} \\
\bottomrule
\end{tabular}

\vspace{-12pt}

\end{table}

\parag{Are the gains due to synthetic data alone?}
To isolate the effect of the training signal from that of the synthetic data,  we compare different supervision signals in \autoref{tab:training_signal},
keeping the base model, the synthetic scenes, and the number of training steps
fixed. 
SFT and GRPO~\citep{shao2024deepseekmath} yield only marginal or no gains, while OPSD achieves the highest average of the three.
These results show that synthetically generated scenes with spatially grounded hints yield consistent improvements over a range of visual skills when used in the OPSD training. We provide more details on different training configurations in  \autoref{sup:baselines} of the Appendix.

\parag{Does spatial guidance matter?}
The choice of privileged information substantially affects performance (\autoref{tab:training_signal}). Without privileged information, OPSD remains at the base-model average, highlighting the importance of teacher--student information asymmetry. Providing the correct answer improves performance, but spatial coordinates alone, without the final count, yield a higher average (72.35 vs.\ 71.27). Combining coordinates with the count achieves the highest average (72.94). In contrast, providing cropped images of the target objects, as in Vision-OPD, reduces the average and substantially degrades CountQA. These ablations highlight the importance of spatially grounded guidance, beyond simply revealing the correct answer or providing localized visual observations.

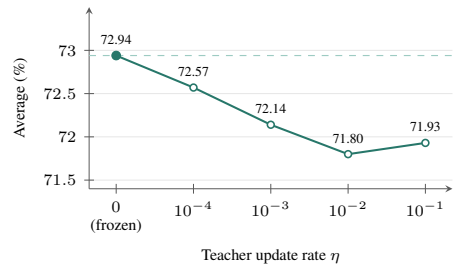
\begin{wrapfigure}{r}{0.43\linewidth}
\vspace{-12pt}
\centering
\resizebox{\linewidth}{!}{\colorlet{ink}{black}%
\colorlet{line}{PineGreen!70!black}%
\colorlet{ours}{PineGreen!70!black}%
\colorlet{rule}{PineGreen!35}%
\colorlet{axis}{black!65}%

\newcommand{\plotw}{56mm}%
\newcommand{\ploth}{28mm}%

\begin{tikzpicture}[font=\scriptsize]
\begin{axis}[
  width=\plotw, height=\ploth, scale only axis,
  xlabel={Teacher update rate $\eta$},
  ylabel={Average (\%)},
  every axis label/.style={font=\scriptsize, text=ink},
  xtick={0,1,2,3,4}, xmin=-0.35, xmax=4.35,
  xticklabels={$0$\\(frozen), $10^{-4}$, $10^{-3}$, $10^{-2}$, $10^{-1}$},
  xticklabel style={font=\scriptsize, text=ink, align=center},
  ymin=71.4, ymax=73.5, ytick={71.5, 72, 72.5, 73},
  yticklabel style={font=\scriptsize, text=ink},
  axis lines=left, axis line style={draw=axis, line width=0.4pt},
  tick style={draw=axis, line width=0.4pt},
  ymajorgrids, grid style={draw=black!10, line width=0.3pt},
  clip=false,
  nodes near coords, point meta=explicit symbolic,
  every node near coord/.style={font=\tiny, text=ink, anchor=south,
                                yshift=1.2pt},
]

\addplot[draw=rule, line width=0.5pt, dashed, forget plot, no markers]
  coordinates {(-0.35,72.94) (4.35,72.94)};

\addplot[draw=line, line width=0.9pt, mark=*, mark size=1.5pt,
         mark options={fill=white, draw=line, line width=0.7pt}]
  coordinates {
    (0, 72.94) [72.94]
    (1, 72.57) [72.57]
    (2, 72.14) [72.14]
    (3, 71.80) [71.80]
    (4, 71.93) [71.93]
  };

\addplot[only marks, mark=*, mark size=1.9pt, mark options={fill=ours, draw=ours},
         forget plot] coordinates {(0,72.94)};

\end{axis}
\end{tikzpicture}}
\vspace{-16pt}
\caption{\textbf{Impact of the teacher update rate on Qwen3.5-4B} when trained with \ours. We report the average performance across benchmarks against the EMA rate $\eta$; the filled point and dashed line mark training with teacher frozen.
}
\label{fig:teacher_rate}
\vspace{-10pt}
\end{wrapfigure}

\parag{How much synthetic data is needed?}
The data-size ablations in ~\autoref{app:results} (\autoref{tab:data_size}) show that, with one epoch per setting, $3{,}000$ scenes is sufficient yielding the best performance
for both MLLMs among the tested data sizes.
Additional scene-generation ablations in ~\autoref{app:results} (\autoref{tab:scene_generation}) show that our default $1024 \times 1024$ resolution and $M \in \{12, \dots, 40\}$ objects per scene give the highest average among the tested settings, although we notice that CountQA benefits from denser scenes.

\parag{Frozen or EMA teacher?}
In \autoref{fig:teacher_rate}, we examine the impact of updating the teacher during training. Interestingly, keeping the teacher frozen ($\eta = 0$) yields the best average performance, and increasing $\eta$ generally degrades it. We hypothesize that updating the teacher on synthetic images may introduce a slight domain shift, diminishing the overall gains. Detailed per-benchmark results are provided in Appendix~\autoref{tab:teacher_rate}.

\begin{table}[t]
\caption{\textbf{Best tested inference settings on Qwen3.5-4B.}
For each model and benchmark, the reported accuracy (\%) is the best
across thinking enabled or thinking disabled. Settings are
selected separately for each cell; Avg. is the mean of the 15 selected
scores. Smaller numbers show percentage-point changes from the base
model. CVB: CVBench; MME-RW:
MME-RealWorld; OCRB: OCRBench; HallB: HallusionBench; HR-4K/8K:
HR-Bench 4K/8K; Zoom: ZoomBench.}
\label{tab:thinking}
\centering
\scriptsize
\setlength{\tabcolsep}{1.1pt}
\renewcommand{\arraystretch}{1.15}
\begin{tabular}{l ccccccccccc !{\hspace{5pt}} cccc !{\hspace{5pt}} >{\columncolor{avgcol}}c}
\toprule
 & \multicolumn{11}{c}{General} & \multicolumn{4}{c}{High-resolution} & \multicolumn{1}{c}{} \\
\cmidrule(lr){2-12} \cmidrule(lr){13-16}
Model & \hd{CVB} & \hd{MME-RW} & \hd{BLINK} & \hd{GQA} & \hd{CountQA} & \hd{ChartQA} & \hd{DocVQA} & \hd{OCRB} & \hd{EvoChart} & \hd{HallB} & \hd{AMBER} & \hd{V$^*$} & \hd{HR-4K} & \hd{HR-8K} & \hd{Zoom} & \hd{Avg.} \\
\midrule
\rowcolor{baserow} Qwen3.5-4B & 89.16 & 70.30 & \textbf{71.23} & 69.32 & 46.86 & 86.28 & 96.45 & 87.30 & 77.36 & 69.82 & 88.80 & 82.72 & \textbf{88.12} & 81.50 & 50.65 & 77.06 \\
Vision-OPD & \dc{88.51}{\dd{0.65}} & \dc{\textbf{75.26}}{\du{4.96}} & \dc{67.75}{\dd{3.48}} & \dc{68.23}{\dd{1.09}} & \dc{42.74}{\dd{4.12}} & \dc{86.04}{\dd{0.24}} & \dc{96.24}{\dd{0.21}} & \dc{87.00}{\dd{0.30}} & \dc{75.52}{\dd{1.84}} & \dc{67.09}{\dd{2.73}} & \dc{86.32}{\dd{2.48}} & \dc{\textbf{92.15}}{\du{9.43}} & \dc{83.50}{\dd{4.62}} & \dc{81.00}{\dd{0.50}} & \dc{\textbf{65.21}}{\du{14.56}} & \dc{77.50}{\du{0.45}} \\[2pt]
\rowcolor{oursrow} \ours & \dc{\textbf{90.45}}{\du{1.29}} & \dc{74.59}{\du{4.29}} & \dc{\textbf{71.23}}{\de{0.00}} & \dc{\textbf{75.39}}{\du{6.07}} & \dc{\textbf{54.12}}{\du{7.26}} & \dc{\textbf{86.56}}{\du{0.28}} & \dc{\textbf{97.01}}{\du{0.56}} & \dc{\textbf{89.20}}{\du{1.90}} & \dc{\textbf{78.64}}{\du{1.28}} & \dc{\textbf{81.60}}{\du{11.78}} & \dc{\textbf{94.37}}{\du{5.57}} & \dc{88.48}{\du{5.76}} & \dc{86.62}{\dd{1.50}} & \dc{\textbf{83.25}}{\du{1.75}} & \dc{53.61}{\du{2.96}} & \dc{\textbf{80.34}}{\du{3.28}} \\[2pt]
\bottomrule
\end{tabular}

\vspace{-12pt}
\end{table}

\parag{Effect of thinking mode.} 
For computational efficiency, all models are trained and, by default, evaluated with thinking disabled. To study the effect of additional inference-time compute, in \autoref{tab:thinking} we report, the better of the two results obtained with thinking enabled and disabled for all 15 benchmarks.
Comparing the baseline model with \ours{}, we find that \ours{} improves on average by 3.28 and on majority of benchmarks. Vision-OPD yields only a modest average gain of 0.45 points: its improvements are concentrated on V*, ZoomBench, and MME-RealWorld, while it decreases accuracy on the remaining 12 benchmarks. Additionally, we provide the results with thinking always enabled in Appendix \autoref{tab:thinking_only}.

\begin{figure}[t]
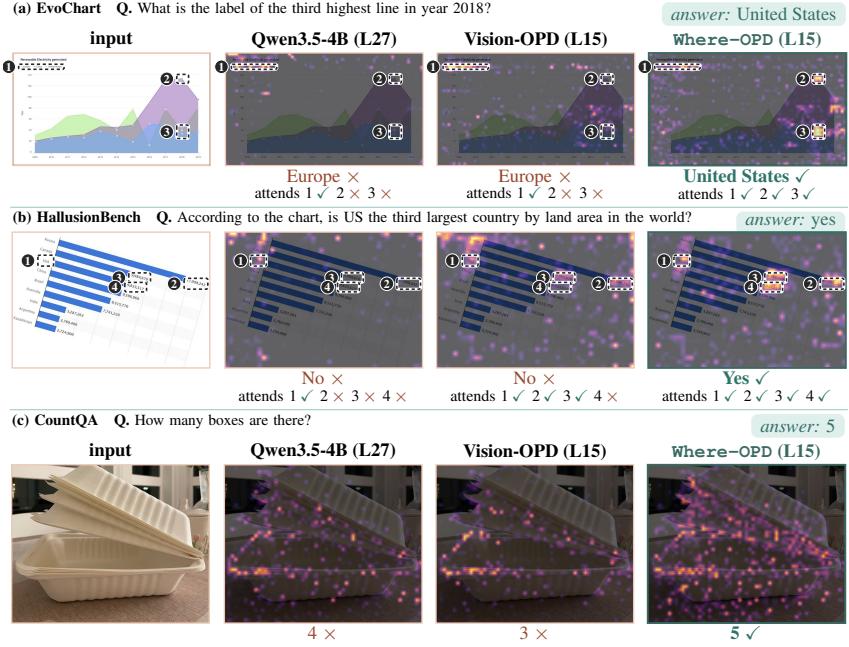

\centering
\IfFileExists{ve_common.tex}{\input{ve_common}}{\input{figures/ve_common}}%
\graphicspath{{images/}{figures/images/}}%
\renewcommand{\vencols}{3}\renewcommand{\panw}{26mm}%
\renewcommand{\inw}{\panw}%
\renewcommand{\qw}{80mm}%

\begin{tikzpicture}[ve, q/.append style={font=\fontsize{5.8}{7}\selectfont}]

\renewcommand{\veprefix}{\textbf{(a)~EvoChart}\quad}%
\veblock[4.8mm]{a}{(0,0)}
  {What is the label of the third highest line in year 2018?}
  {United States}{evochart_000047}
  {base/Europe/0/{attends \rin{1} \rout{2} \rout{3}}, visionopd/Europe/0/{attends \rin{1} \rout{2} \rout{3}}, ours/United States/1/{attends \rin{1} \rin{2} \rin{3}}}
\veheads{a}{Qwen3.5-4B (L27), Vision-OPD (L15), \ours (L15)}
\foreach \pnl in {ain,ap1,ap2,ap3}{%
  \veregions{\pnl}{0.03/0.10/0.26/0.145/1, 0.83/0.19/0.89/0.27/2, 0.83/0.64/0.89/0.76/3}}

\vesep{a}
\renewcommand{\veprefix}{\textbf{(b)~HallusionBench}\quad}%
\renewcommand{\qw}{96mm}%
\veblock{b}{\venext{a}}
  {According to the chart, is US the third largest country by land area in the world?}
  {yes}{hallusion_000033}
  {base/No/0/{attends \rin{1} \rout{2} \rout{3} \rout{4}}, visionopd/No/0/{attends \rin{1} \rin{2} \rin{3} \rout{4}}, ours/Yes/1/{attends \rin{1} \rin{2} \rin{3} \rin{4}}}
\foreach \pnl in {bin,bp1,bp2,bp3}{%
  \veregions{\pnl}{0.13/0.17/0.21/0.25/1, 0.87/0.34/0.99/0.42/2, 0.59/0.30/0.70/0.37/3, 0.57/0.37/0.68/0.44/4}}

\vesep{b}
\renewcommand{\qw}{80mm}%
\renewcommand{\veprefix}{\textbf{(c)~CountQA}\quad}%
\veblock[4.8mm]{c}{\venext{b}}
  {How many boxes are there?}{5}{countqa_000176}
  {base/4/0/, visionopd/3/0/, ours/5/1/}
\veheads{c}{Qwen3.5-4B (L27), Vision-OPD (L15), \ours (L15)}

\pgfresetboundingbox
\useasboundingbox (abg.north west) rectangle (cbg.south east);

\end{tikzpicture}
\vspace{-10pt}
\caption{\textbf{Visual evidence on EvoChart, HallusionBench and CountQA.}
We show each visual input, question, ground-truth answer, and attention maps from the response tokens to the image for Qwen3.5-4B, Vision-OPD, and \ours. Layers are selected on held-out synthetic scenes (\autoref{tab:layer_selection}); predictions appear below each map.
Numbered dashed boxes indicate question-relevant regions, e.g., the legend and 2018 values in (a), and the USA label and land areas of Russia, Canada, and the USA in (b), added solely for visualization and absent from the models' inputs. Checkmarks indicate regions receiving at least twice their area-proportional share of attention.
Unlike the baselines, \ours attends to all marked regions in (a) and (b), suggesting better alignment with question-relevant visual evidence. In (c), its attention more clearly follows the boundaries of the stacked boxes, correctly predicting five where the baselines fail.}
\label{fig:visual_evidence}
\vspace{-12pt}

\end{figure}

\parag{Qualitative examples: Attention to question-relevant visual evidence.}
\autoref{fig:visual_evidence} presents qualitative examples of Qwen3.5-4B post-trained with \ours. 
On EvoChart and HallusionBench, \ours not only predicts the correct answers but also attends to all marked question-relevant regions, unlike the base model and Vision-OPD. 
On CountQA (\autoref{fig:visual_evidence}c),
\ours exhibits attention that more clearly follows the boundaries of the stacked boxes and correctly counts all five instances, whereas both baselines fail. These examples suggest that spatially grounded guidance encourages better alignment of attention with the visual evidence needed to answer each question.

\section{Conclusion}
We introduce \ours, which uses spatially grounded guidance as privileged information for on-policy self-distillation of multimodal large language models.
Using procedurally generated scenes, we provide the teacher with textual hints identifying question-relevant elements and their locations, while the student learns from the image and question alone.
Compared with methods that give the teacher enhanced visual inputs, \ours achieves improvements across a broader range of evaluated tasks, transferring to real-world images despite training exclusively on synthetic scenes.
Ablations show that spatial grounding is important: 
removing the hint or providing the answer without locations yields lower performance. Together, these results suggest that privilege in the form of guidance towards relevant visual evidence provides a transferable distillation signal.

\section*{Acknowledgments}
This work was supported by the European Union’s Horizon Europe research and innovation programme under grant agreement number 101214398 (ELLIOT), by HPC resources from GENCI-IDRIS (Grants AD011015037R2,  A0201016980), and project RODEO (ANR-24-CE23-5886). We thank Yijiang Li for sharing checkpoints for S$^{2}$VOPD and Yishuo Cai for sharing checkpoints for Imagine-OPD.

\bibliography{iclr2027_conference}
\bibliographystyle{iclr2027_conference}

\appendix
\clearpage
\begin{center}
{\Large\scshape Appendix}
\end{center}
\startcontents[appendix]
\printcontents[appendix]{l}{1}{\setcounter{tocdepth}{2}}
\vspace{1em}

\section{Implementation details}
\label{sec:training-details}

\subsection{Synthetic counting data}
\label{sec:supp:synthdata}

All training sets are generated procedurally.

Each scene is a $1024 \times 1024$ frame with a solid-color background. The background color is sampled from ten base colors—seven light neutrals and tints and three dark colors—with small per-channel jitter, so exact background colors are rarely repeated. Each object combines one of twelve colors with one of seven shapes (circle, square, triangle, diamond, star, cross, or hexagon), with each shape parameterized by a single radius. 

For each scene, we first sample a small vocabulary of two to five colors and two to four shapes. We then place $12$–$40$ objects, drawing each object’s shape and color from that vocabulary, its radius from $14$–$64$ pixels, and its center within the frame, rejecting any placement that overlaps an existing object. All scene-generation draws are uniform. The question targets a color–shape pair present in the scene, selected with probability proportional to its count so that questions favor pairs with more objects over singletons. The multiple-choice distractors are near misses of the true count. The resulting training set of $3{,}000$ scenes contains a median of $26$ objects per scene, and the target count ranges from $1$ to $14$, with a median of $3$.

\begin{figure}[t]
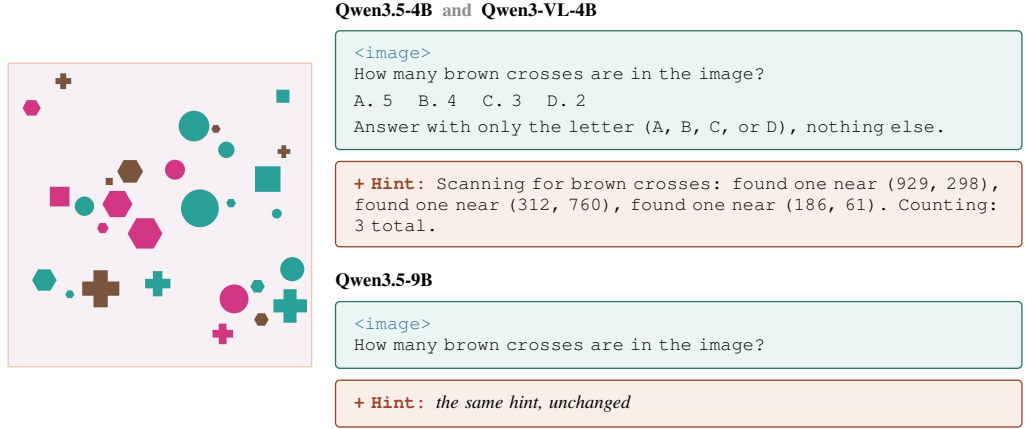

\centering
\includestandalone{figures/train_input_tikz}
\caption{\textbf{Training inputs across MLLMs.}
All three MLLMs use the same generated scenes. Qwen3.5-4B and
Qwen3-VL-4B receive a multiple-choice question, while Qwen3.5-9B
receives an open-ended version. The student receives the image and
question (green box); the teacher receives the same input with the
grounded hint appended (red box).}
\label{fig:train_input}
\end{figure}

Because the generator knows every object’s position, the teacher’s hint is generated directly from the scene state and is exact by construction, requiring no manual annotation. For the main models, the hint narrates a search over the target coordinates and concludes with the count. It therefore provides three pieces of information at once: the target object type, the locations of its instances, and the answer. \autoref{fig:train_input} shows one record exactly as each model receives it. 

For the 4B models, we used multiple-choice questions, whereas for the 9B model, we used open-ended questions (see \autoref{fig:train_input}). This choice was motivated by our observation that multiple-choice questions frequently elicited direct answers without intermediate reasoning tokens from the 9B model, limiting the supervision available for on-policy self-distillation.

\subsection{Training details}
\label{sup:training}

We train all of the models for one epoch with eight on-policy rollouts per prompt, a global batch size of 96, and a learning rate of $2\times10^{-6}$ with 10 warm-up steps. Rollouts are generated with vLLM using stochastic sampling, with thinking disabled during training. For Qwen3.5-9B, we use the same optimization setup with a global batch size of 48. We additionally evaluate transfer to Qwen3-VL-4B-Instruct using the same synthetic supervision construction. Training \ours on the $3{,}000$ set takes about 1.4 h for Qwen3.5-4B and 1.2 h for Qwen3-VL-4B, and about 5.8 h for Qwen3.5-9B, all on 4 H100 GPUs.

\subsection{Evaluation Protocol}
 All models are evaluated with the same pipeline. Responses are generated using greedy decoding, with thinking disabled. Each response is then scored in three stages. First, a rule-based verifier extracts and checks the final answer. Second, for multiple-choice questions, the option letter that the response concludes with is compared with the ground truth. Third, any response that neither stage resolves is passed to an LLM judge, which compares the response with the reference answer and returns a binary verdict. Accuracy is the fraction of questions judged correct, with unparseable or failed responses counted as incorrect. The judge and the scoring rules are identical across all methods.

\subsection{Baselines: GRPO and supervised fine-tuning}
\label{sup:baselines}

Both baselines use the same $3{,}000$ examples that were used to train \ours. Each is trained for one epoch with a global batch of $96$.
\paragraph{GRPO with a verifiable reward.} We sample $8$ rollouts per sample and assign a binary reward by comparing the extracted option letter with the ground truth. Advantages are centered by the group mean. Learning rate set to $2\times10^{-6}$.
\paragraph{Supervised fine-tuning.} We minimize cross-entropy on the grounded coordinate hint as well as the ground truth final count, using  learning rate of $1\times10^{-5}$.

\section{Additional Results}
\label{app:results}

\subsection{Are results driven by in domain tasks?} In \autoref{tab:per_category} we show per category results for 6 benchmarks. We examine whether the benchmark-level gains are driven mainly by counting questions, the task seen during training. Improvements span across diverse visual categories.
\begin{table}[t]
\caption{\textbf{Per-category accuracy on Qwen3.5-4B.}
Accuracy (\%) for Qwen3.5-4B and \ours across six benchmarks,
broken down by their category labels. $n$ is the
number of examples in each category, and $\Delta$ is \ours minus base
accuracy. Bold indicates the higher score.}
\label{tab:per_category}
\centering
\scriptsize
\setlength{\tabcolsep}{1.8pt}
\renewcommand{\arraystretch}{1.1}
\begin{minipage}[t]{0.47\linewidth}\centering
\begin{tabular}{l r r r r}
\toprule
Category & $n$ & Qwen3.5-4B & \ours & $\Delta$\\
\midrule
\addlinespace[1pt]\multicolumn{5}{l}{\textbf{CVBench}}\\[1pt]
Count & 788 & 72.72 & \textbf{76.65} & \textcolor{deltaup}{+3.93}\\
Depth & 600 & 95.50 & \textbf{96.33} & \textcolor{deltaup}{+0.83}\\
Distance & 600 & 90.17 & \textbf{93.50} & \textcolor{deltaup}{+3.33}\\
Relation & 650 & 95.38 & \textbf{96.77} & \textcolor{deltaup}{+1.39}\\
\emph{Overall} & 2,638 & 87.45 & \textbf{89.92} & \textcolor{deltaup}{+2.47}\\
\addlinespace[1pt]\multicolumn{5}{l}{\textbf{V$^*$}}\\[1pt]
direct attributes & 115 & 81.74 & \textbf{89.57} & \textcolor{deltaup}{+7.83}\\
relative position & 76 & 84.21 & \textbf{86.84} & \textcolor{deltaup}{+2.63}\\
\emph{Overall} & 191 & 82.72 & \textbf{88.48} & \textcolor{deltaup}{+5.76}\\
\addlinespace[1pt]\multicolumn{5}{l}{\textbf{BLINK}}\\[1pt]
Art style & 117 & 76.07 & \textbf{79.49} & \textcolor{deltaup}{+3.42}\\
Counting & 120 & 71.67 & \textbf{80.83} & \textcolor{deltaup}{+9.16}\\
Forensic detection & 132 & 57.58 & \textbf{62.12} & \textcolor{deltaup}{+4.54}\\
Functional corresp. & 130 & \textbf{56.92} & 55.38 & \textcolor{deltadn}{$-$1.54}\\
IQ test & 150 & 20.67 & \textbf{28.67} & \textcolor{deltaup}{+8.00}\\
Jigsaw & 150 & \textbf{64.00} & 63.33 & \textcolor{deltadn}{$-$0.67}\\
Multi-view reasoning & 133 & \textbf{48.12} & 42.11 & \textcolor{deltadn}{$-$6.01}\\
Object localization & 122 & 53.28 & \textbf{56.56} & \textcolor{deltaup}{+3.28}\\
Relative depth & 124 & 82.26 & 82.26 & \textcolor{black!45}{$\pm$0.00}\\
Relative reflectance & 134 & 67.16 & 67.16 & \textcolor{black!45}{$\pm$0.00}\\
Semantic corresp. & 139 & \textbf{66.91} & 65.47 & \textcolor{deltadn}{$-$1.44}\\
Spatial relation & 143 & \textbf{88.11} & 87.41 & \textcolor{deltadn}{$-$0.70}\\
Visual corresp. & 172 & 91.28 & 91.28 & \textcolor{black!45}{$\pm$0.00}\\
Visual similarity & 135 & 82.22 & \textbf{88.89} & \textcolor{deltaup}{+6.67}\\
\emph{Overall} & 1,901 & 66.28 & \textbf{67.96} & \textcolor{deltaup}{+1.68}\\
\bottomrule
\end{tabular}
\end{minipage}\hfill
\begin{minipage}[t]{0.52\linewidth}\centering
\begin{tabular}{l r r r r}
\toprule
Category & $n$ & Qwen3.5-4B & \ours & $\Delta$\\
\midrule
\addlinespace[1pt]\multicolumn{5}{l}{\textbf{MME-RealWorld}}\\[1pt]
P: autonomous driving & 3,660 & 33.06 & \textbf{57.76} & \textcolor{deltaup}{+24.70}\\
P: diagram and table & 5,433 & 89.32 & \textbf{90.45} & \textcolor{deltaup}{+1.13}\\
P: monitoring & 2,196 & 42.94 & \textbf{51.68} & \textcolor{deltaup}{+8.74}\\
P: OCR complex context & 5,740 & 82.44 & \textbf{92.30} & \textcolor{deltaup}{+9.86}\\
P: remote sensing & 3,738 & 56.05 & \textbf{63.88} & \textcolor{deltaup}{+7.83}\\
R: autonomous driving & 1,344 & 28.79 & \textbf{51.56} & \textcolor{deltaup}{+22.77}\\
R: diagram and table & 500 & 84.40 & \textbf{85.40} & \textcolor{deltaup}{+1.00}\\
R: monitoring & 498 & 34.34 & \textbf{50.20} & \textcolor{deltaup}{+15.86}\\
R: OCR complex context & 500 & 64.40 & \textbf{78.20} & \textcolor{deltaup}{+13.80}\\
\emph{Overall} & 23,609 & 64.11 & \textbf{74.59} & \textcolor{deltaup}{+10.48}\\
\addlinespace[1pt]\multicolumn{5}{l}{\textbf{HallusionBench}}\\[1pt]
chart & 130 & 79.23 & \textbf{90.77} & \textcolor{deltaup}{+11.54}\\
figure & 80 & 81.25 & \textbf{85.00} & \textcolor{deltaup}{+3.75}\\
illusion & 144 & 62.50 & \textbf{74.31} & \textcolor{deltaup}{+11.81}\\
map & 64 & 70.31 & \textbf{78.12} & \textcolor{deltaup}{+7.81}\\
math & 108 & 53.70 & \textbf{81.48} & \textcolor{deltaup}{+27.78}\\
ocr & 143 & 79.72 & \textbf{85.31} & \textcolor{deltaup}{+5.59}\\
table & 112 & 79.46 & \textbf{91.96} & \textcolor{deltaup}{+12.50}\\
video & 170 & 56.47 & \textbf{70.59} & \textcolor{deltaup}{+14.12}\\
\emph{Overall} & 951 & 69.40 & \textbf{81.60} & \textcolor{deltaup}{+12.20}\\
\addlinespace[1pt]\multicolumn{5}{l}{\textbf{AMBER}}\\[1pt]
attribute & 7,628 & 85.55 & \textbf{93.63} & \textcolor{deltaup}{+8.08}\\
existence & 4,924 & 95.43 & \textbf{96.24} & \textcolor{deltaup}{+0.81}\\
relation & 1,664 & 84.01 & \textbf{92.19} & \textcolor{deltaup}{+8.18}\\
\emph{Overall} & 14,216 & 88.80 & \textbf{94.37} & \textcolor{deltaup}{+5.57}\\
\bottomrule
\end{tabular}
\end{minipage}

\end{table}

\subsection{Scene-generation and training ablations}

\paragraph{Scene-generation} We vary the resolution and number of objects in the synthetic training scenes (\autoref{tab:scene_generation}). The default $1024^2$ resolution and $12$--$40$ objects per scene achieve the highest seven-benchmark average among the tested settings. 
\begin{table}[t]
\caption{\textbf{Scene-generation ablations on Qwen3.5-4B.}
Top: the same scenes and questions rendered at different resolutions,
with hint coordinates rescaled accordingly. Bottom: different ranges
for the number of objects per scene. Results are accuracy (\%);
Avg. is the unweighted mean across seven benchmarks. The highlighted
rows show the default setting in both blocks; bold marks the best
result within each block. CVB: CVBench; HR-4K/8K: HR-Bench 4K/8K;
Zoom: ZoomBench.}
\label{tab:scene_generation}
\centering
\small
\setlength{\tabcolsep}{4pt}
\renewcommand{\arraystretch}{1.15}
\begin{tabular}{l ccccccc !{\hspace{5pt}} >{\columncolor{avgcol}}c}
\toprule
 & \multicolumn{7}{c}{Benchmarks} & \multicolumn{1}{c}{} \\
\cmidrule(lr){2-8}
 & CVB & V$^*$ & HR-4K & HR-8K & Zoom & CountQA & EvoChart & Avg. \\
\midrule
\rowcolor{baserow} \bb{Qwen3.5-4B} & 87.45 & 82.72 & 86.25 & 80.00 & 50.65 & 32.00 & 68.32 & 69.63 \\
\midrule
\multicolumn{9}{@{}l}{\emph{Training-scene resolution}} \\
\rowcolor{oursrow} $1024^2$ (ours) & 89.92 & \textbf{88.48} & \textbf{85.50} & \textbf{83.25} & \textbf{51.01} & 33.77 & \textbf{78.64} & \textbf{72.94} \\
$768^2$ & \textbf{89.99} & 86.39 & 85.12 & 79.50 & 49.94 & 33.57 & 77.52 & 71.72 \\
$512^2$ & 88.78 & 85.86 & 84.12 & 81.12 & 50.41 & \textbf{34.69} & \textbf{78.64} & 71.95 \\
\midrule
\multicolumn{9}{@{}l}{\emph{Objects per scene}} \\
\rowcolor{oursrow} 12--40 (ours) & \textbf{89.92} & \textbf{88.48} & 85.50 & \textbf{83.25} & \textbf{51.01} & 33.77 & \textbf{78.64} & \textbf{72.94} \\
25--100 & 89.16 & 83.25 & 85.62 & 80.88 & 49.94 & 34.62 & 77.44 & 71.56 \\
50--200 & 88.82 & 82.20 & \textbf{86.12} & 82.38 & 48.99 & \textbf{35.41} & 76.96 & 71.55 \\
\bottomrule
\end{tabular}

\end{table}

\paragraph{Training-set size}
By training one epoch on each setting, \autoref{tab:data_size} shows that the benchmark average is highest at $3{,}000$ scenes for both MLLMs among the sizes tested. Larger training sets improve some individual benchmarks but do not improve the average. We therefore use $3{,}000$ scenes in the main experiments.
\begin{table}[t]
\caption{\textbf{Training-set size.}
Accuracy (\%) for Qwen3.5-4B (top) and Qwen3-VL-4B (bottom) after one epoch
at each dataset size. The
highlighted rows are the setting used in the main experiments. Avg. is the
unweighted mean across seven benchmarks; bold marks the best dataset size
within each backbone. CVB: CVBench; HR-4K/8K: HR-Bench 4K/8K; Zoom: ZoomBench.}
\label{tab:data_size}
\centering
\small
\setlength{\tabcolsep}{4pt}
\renewcommand{\arraystretch}{1.15}
\begin{tabular}{r r ccccccc !{\hspace{5pt}} >{\columncolor{avgcol}}c}
\toprule
 &  & \multicolumn{7}{c}{Benchmarks} & \multicolumn{1}{c}{} \\
\cmidrule(lr){3-9}
Examples & Steps & CVB & V$^*$ & HR-4K & HR-8K & Zoom & CountQA & EvoChart & Avg. \\
\midrule
\rowcolor{baserow} \multicolumn{2}{l}{Qwen3.5-4B} & 87.45 & 82.72 & 86.25 & 80.00 & 50.65 & 32.00 & 68.32 & 69.63 \\
1,000 & 10 & 87.00 & 84.29 & 83.50 & 79.00 & 47.93 & 34.42 & 70.24 & 69.48 \\
2,000 & 20 & 89.54 & 86.91 & \textbf{85.88} & 81.88 & 50.65 & 33.97 & 78.24 & 72.44 \\
\rowcolor{oursrow} 3,000 & 31 & \textbf{89.92} & \textbf{88.48} & 85.50 & \textbf{83.25} & \textbf{51.01} & 33.77 & 78.64 & \textbf{72.94} \\
6,000 & 62 & 88.97 & 83.77 & 84.12 & 81.50 & 49.94 & 35.86 & \textbf{79.20} & 71.91 \\
11,000 & 114 & 89.50 & 86.39 & 83.38 & 79.62 & 50.06 & \textbf{36.26} & 78.00 & 71.89 \\
\midrule
\rowcolor{baserow} \multicolumn{2}{l}{Qwen3-VL-4B} & 85.44 & 81.15 & 80.00 & 75.12 & 45.09 & 27.68 & 62.48 & 65.28 \\
1,000 & 10 & 85.03 & 82.20 & 79.00 & 74.12 & 44.62 & 29.12 & 63.04 & 65.30 \\
\rowcolor{oursrow} 3,000 & 31 & 86.58 & \textbf{84.29} & \textbf{81.38} & 75.25 & 45.92 & \textbf{30.37} & \textbf{64.24} & \textbf{66.86} \\
6,000 & 62 & 86.66 & 83.25 & 80.88 & \textbf{75.88} & 45.80 & \textbf{30.37} & 64.16 & 66.71 \\
11,000 & 114 & \textbf{87.23} & 83.25 & 77.88 & 75.50 & \textbf{47.22} & 23.63 & 63.36 & 65.44 \\
\bottomrule
\end{tabular}

\label{tab:data_size_vl}
\end{table}

\paragraph{EMA update rate}In \autoref{tab:teacher_rate}, we report per-benchmark accuracy at different teacher update rates for Qwen3.5-4B and Qwen3-VL-4B.
\begin{table}[t]
\caption{\textbf{Teacher update-rate study.}
Accuracy (\%) for Qwen3.5-4B (top) and Qwen3-VL-4B (bottom) at
different EMA update rates $\eta$; $\eta=0$ leaves the teacher frozen.
The highlighted rows
are used in the main experiments. Avg. is the unweighted mean across
seven benchmarks; bold marks the best update rate within each
backbone. CVB: CVBench; HR-4K/8K: HR-Bench 4K/8K; Zoom: ZoomBench.}
\label{tab:teacher_rate}
\centering
\small
\setlength{\tabcolsep}{4pt}
\renewcommand{\arraystretch}{1.15}
\begin{tabular}{l ccccccc !{\hspace{5pt}} >{\columncolor{avgcol}}c}
\toprule
 & \multicolumn{7}{c}{Benchmarks} & \multicolumn{1}{c}{} \\
\cmidrule(lr){2-8}
Teacher update rate $\eta$ & CVB & V$^*$ & HR-4K & HR-8K & Zoom & CountQA & EvoChart & Avg. \\
\midrule
\rowcolor{baserow} Qwen3.5-4B & 87.45 & 82.72 & 86.25 & 80.00 & 50.65 & 32.00 & 68.32 & 69.63 \\
\rowcolor{oursrow} 0 (frozen) & 89.92 & 88.48 & \textbf{85.50} & \textbf{83.25} & 51.01 & 33.77 & \textbf{78.64} & \textbf{72.94} \\
0.0001 & 89.61 & \textbf{89.01} & 83.12 & 81.50 & \textbf{51.36} & \textbf{35.47} & 77.92 & 72.57 \\
0.001 & 89.31 & 86.91 & 84.75 & 80.38 & 51.01 & 34.75 & 77.84 & 72.14 \\
0.01 & 89.27 & 84.82 & 84.25 & 81.62 & 51.01 & 34.75 & 76.88 & 71.80 \\
0.1 & \textbf{90.30} & 84.82 & 84.75 & 81.12 & 50.06 & 34.49 & 78.00 & 71.93 \\
\midrule
\rowcolor{baserow} Qwen3-VL-4B & 85.44 & 81.15 & 80.00 & 75.12 & 45.09 & 27.68 & 62.48 & 65.28 \\
\rowcolor{oursrow} 0 (frozen) & \textbf{86.58} & 84.29 & \textbf{81.38} & \textbf{75.25} & \textbf{45.92} & \textbf{30.37} & 64.24 & \textbf{66.86} \\
0.0001 & 86.39 & 84.82 & 79.62 & 74.38 & 44.85 & 29.38 & 63.84 & 66.18 \\
0.001 & 86.54 & \textbf{85.34} & 80.25 & 74.75 & 45.21 & 28.80 & 64.08 & 66.42 \\
0.01 & 86.35 & 84.82 & 79.88 & 74.75 & 45.68 & 28.86 & \textbf{64.40} & 66.39 \\
0.1 & 86.32 & \textbf{85.34} & 79.50 & 75.00 & 45.09 & 30.10 & 63.84 & 66.46 \\
\bottomrule
\end{tabular}
\label{tab:teacher_rate_vl}
\end{table}

\paragraph{Thinking enabled} In \autoref{tab:thinking_only}, we report accuracy for \ours relative to base model Qwen3.5-4B across all 15 benchmarks with thinking enabled.
\begin{table}[t]
\caption{\textbf{Evaluations under thinking enabled mode on Qwen3.5-4B.}
All models are evaluated with thinking enabled. Results are accuracy
(\%); Avg. is the unweighted mean across 15 benchmarks. The smaller
numbers below each score show the percentage-point change from the
base model evaluated in the same mode. CVB: CVBench; MME-RW:
MME-RealWorld; OCRB: OCRBench; HallB: HallusionBench; HR-4K/8K:
HR-Bench 4K/8K; Zoom: ZoomBench.}
\label{tab:thinking_only}
\centering
\scriptsize
\setlength{\tabcolsep}{1.1pt}
\renewcommand{\arraystretch}{1.15}
\begin{tabular}{l ccccccccccc !{\hspace{5pt}} cccc !{\hspace{5pt}} >{\columncolor{avgcol}}c}
\toprule
 & \multicolumn{11}{c}{General} & \multicolumn{4}{c}{High-resolution} & \multicolumn{1}{c}{} \\
\cmidrule(lr){2-12} \cmidrule(lr){13-16}
Model & \hd{CVB} & \hd{MME-RW} & \hd{BLINK} & \hd{GQA} & \hd{CountQA} & \hd{ChartQA} & \hd{DocVQA} & \hd{OCRB} & \hd{EvoChart} & \hd{HallB} & \hd{AMBER} & \hd{V$^*$} & \hd{HR-4K} & \hd{HR-8K} & \hd{Zoom} & \hd{Avg.} \\
\midrule
\rowcolor{baserow} \bb{Qwen3.5-4B} & 89.16 & 70.30 & \textbf{71.23} & 64.86 & 46.86 & \textbf{86.28} & 96.45 & \textbf{86.90} & \textbf{77.36} & \textbf{69.82} & \textbf{88.61} & 82.72 & \textbf{88.12} & 81.50 & 46.15 & 76.42 \\[2pt]
Vision-OPD & \dc{88.51}{\dd{0.65}} & \dc{70.43}{\du{0.13}} & \dc{67.75}{\dd{3.48}} & \dc{62.26}{\dd{2.60}} & \dc{42.74}{\dd{4.12}} & \dc{86.04}{\dd{0.24}} & \dc{96.24}{\dd{0.21}} & \dc{85.80}{\dd{1.10}} & \dc{75.52}{\dd{1.84}} & \dc{67.09}{\dd{2.73}} & \dc{83.88}{\dd{4.73}} & \dc{\textbf{92.15}}{\du{9.43}} & \dc{83.50}{\dd{4.62}} & \dc{81.00}{\dd{0.50}} & \dc{\textbf{57.99}}{\du{11.84}} & \dc{76.06}{\dd{0.36}} \\[2pt]
OPD-V & \dc{86.35}{\dd{2.81}} & \dc{69.81}{\dd{0.49}} & \dc{66.12}{\dd{5.11}} & \dc{58.45}{\dd{6.41}} & \dc{45.03}{\dd{1.83}} & \dc{84.04}{\dd{2.24}} & \dc{96.09}{\dd{0.36}} & \dc{84.70}{\dd{2.20}} & \dc{74.40}{\dd{2.96}} & \dc{64.25}{\dd{5.57}} & \dc{78.13}{\dd{10.48}} & \dc{89.53}{\du{6.81}} & \dc{84.25}{\dd{3.87}} & \dc{79.75}{\dd{1.75}} & \dc{57.28}{\du{11.13}} & \dc{74.55}{\dd{1.88}} \\[2pt]
S2VOPD & \dc{88.82}{\dd{0.34}} & \dc{71.38}{\du{1.08}} & \dc{70.65}{\dd{0.58}} & \dc{61.71}{\dd{3.15}} & \dc{53.93}{\du{7.07}} & \dc{85.04}{\dd{1.24}} & \dc{96.28}{\dd{0.17}} & \dc{86.00}{\dd{0.90}} & \dc{75.20}{\dd{2.16}} & \dc{66.56}{\dd{3.26}} & \dc{83.91}{\dd{4.70}} & \dc{85.34}{\du{2.62}} & \dc{86.75}{\dd{1.37}} & \dc{\textbf{83.50}}{\du{2.00}} & \dc{53.61}{\du{7.46}} & \dc{76.58}{\du{0.16}} \\[2pt]
\rowcolor{oursrow} \ours & \dc{\textbf{90.45}}{\du{1.29}} & \dc{\textbf{71.50}}{\du{1.20}} & \dc{\textbf{71.23}}{\de{0.00}} & \dc{\textbf{65.81}}{\du{0.95}} & \dc{\textbf{54.12}}{\du{7.26}} & \dc{86.16}{\dd{0.12}} & \dc{\textbf{96.49}}{\du{0.04}} & \dc{86.60}{\dd{0.30}} & \dc{76.00}{\dd{1.36}} & \dc{69.40}{\dd{0.42}} & \dc{87.61}{\dd{1.00}} & \dc{83.25}{\du{0.53}} & \dc{86.62}{\dd{1.50}} & \dc{81.75}{\du{0.25}} & \dc{53.61}{\du{7.46}} & \dc{\textbf{77.37}}{\du{0.95}} \\[2pt]
\bottomrule
\end{tabular}

\end{table}

\begin{table}[t]

\centering
\small
\setlength{\tabcolsep}{4pt}
\renewcommand{\arraystretch}{1.15}
\begin{tabular}{lrrrrrrrrc}
\toprule
Model & L3 & L7 & L11 & L15 & L19 & L23 & L27 & L31 & Chosen \\
\midrule
\rowcolor{baserow} Qwen3.5-4B & 0.68 & 0.27 & 0.64 & 0.94 & 0.92 & 0.76 & \textbf{1.15} & 0.54 & L27 \\
Vision-OPD & 1.12 & 0.71 & 3.01 & \textbf{4.75} & 3.20 & 2.31 & 2.57 & 1.47 & L15 \\
\rowcolor{oursrow} \ours & 1.32 & 1.17 & 3.36 & \textbf{5.79} & 4.30 & 2.69 & 2.29 & 1.42 & L15 \\
\bottomrule
\end{tabular}
\caption{\textbf{Attention-layer selection on synthetic counting scenes.}
Mean enrichment of attention on the target objects at each full-attention
layer, measured on a separate selection split of 158 held-out images
with at least three targets. Enrichment of $1$ means the targets receive
attention proportional to their area. \emph{Chosen} is the layer with
the highest enrichment for each model. The base model is Qwen3.5-4B.}
\label{tab:layer_selection}
\end{table}

\subsection{Additional qualitative results}
\begin{figure}[p]
\centering
\IfFileExists{ve_common.tex}{\input{ve_common}}{\input{figures/ve_common}}%
\IfFileExists{tr_common.tex}{\input{tr_common}}{\input{figures/tr_common}}%
\graphicspath{{images/}{figures/images/}}%
\renewcommand{\qw}{36mm}%

\begin{tikzpicture}[ve,
  quote/.style = {font=\scriptsize, align=left, text width=\txw, inner sep=0pt},
]

\renewcommand{\trbench}{CVBench}\renewcommand{\trans}{(A) table}%
\trwide{a}{(0,0)}{loc_c1.jpg}
  {Closer to the lamp (red): the table (blue) or the door (green)?}{table}
  {``The door (green box) is a doorway located on the far wall. The vertical
    alignment of the red and green boxes suggests the lamp is hanging directly above
    or very near the door. \ldots\ the door is physically closer to the lamp than
    the table is.''}{(B) door}
  {``The lamp (red box) is hanging from the ceiling in the middle area of the room.
    The table (blue box) is a coffee table located in the immediate foreground,
    directly beneath the area where the lamp hangs. The door (green box) is located
    further back in the room, near the wall.''}

\trsep{a}
\renewcommand{\trans}{(A) truck}%
\trwide{b}{\trnext{a}}{loc_c2.jpg}
  {Closer to the bus (red): the truck (blue) or the trailer (green)?}{truck}
  {``The bus appears to be roughly parallel with the trailer, perhaps slightly ahead
    or behind, but definitely closer to the trailer than to the truck which is
    further ahead in the left lane.''}{(B) trailer}
  {``The bus is driving in the lane to the right of the truck/trailer. The truck
    (blue box) is positioned ahead of the trailer (green box). \ldots\ The truck is
    adjacent to the bus's lane, while the trailer is further away to the left.''}

\trsep{b}
\renewcommand{\trans}{(B) night stand}%
\trwide{c}{\trnext{b}}{loc_c3.jpg}
  {Closer to the lamp (red): the dresser (blue) or the night stand (green)?}{night stand}
  {``The dresser is located to the left of the bed. It is positioned near the head of
    the bed, adjacent to the lamp's location. \ldots\ The lamp and dresser are
    separated only by the width of the bed, whereas the lamp and the green box object
    are separated by the length of the bed.''}{(A) dresser}
  {``The lamp (red box) is located on the right side of the bed. The night stand
    (green box) is located at the foot of the bed, directly adjacent to where the
    lamp is placed. The dresser (blue box) is located on the far left side of the
    room, separated from the bed and lamp by a significant distance.''}

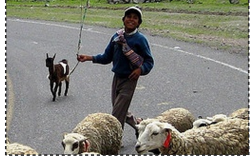
\node[fit=(abg)(bbg)(cbg), inner sep=0pt] (all) {};
\pgfresetboundingbox
\useasboundingbox (all.north west) rectangle (all.south east);

\end{tikzpicture}

\caption{\textbf{Thinking traces on CVBench.}
Three questions comparing distances between objects marked by CVBench's
colored boxes. Both the Qwen3.5-4B base model and ours use thinking
mode. Each row shows the image, question, ground truth answer, and
excerpts from both traces; ellipses mark omitted text. In these
examples, the base model settles the comparison from how the boxes line up in the
frame; \ours appears to place each object in the room before making a prediction.}
\label{fig:traces_cvbench}
\end{figure}

\begin{figure}[p]
\centering
\IfFileExists{ve_common.tex}{\input{ve_common}}{\input{figures/ve_common}}%
\IfFileExists{tr_common.tex}{\input{tr_common}}{\input{figures/tr_common}}%
\graphicspath{{images/}{figures/images/}}%
\renewcommand{\qw}{36mm}%

\begin{tikzpicture}[ve,
  quote/.style = {font=\scriptsize, align=left, text width=\txw, inner sep=0pt},
]

\renewcommand{\trbench}{GQA}\renewcommand{\trans}{goat}%
\renewcommand{\gbb}{}\renewcommand{\gbcap}{base: no boxes}%
\renewcommand{\gbo}{0.290/0.069/0.605/0.845, 0.165/0.354/0.263/0.668}%
\trrow{a}{(0,0)}{tr_goat}{0.55/0.249/0.95/0.599}
  {What is the animal that the person that is walking is leading?}{goat}
  {``The small animal is a dog. It's brown and black, on a leash or lead held by
    the person. \ldots\ Is it possible it's a goat kid? No, looks like a dog.''}{dog}
  {``The adult on the right [666, 273, 792, 545] is holding a stick or rope and is
    walking alongside a small animal. \ldots\ The small animal is a goat
    [616, 373, 655, 483]. It's brown and has horns.''}

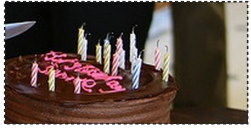
\trsep{a}
\renewcommand{\trbench}{GQA}\renewcommand{\trans}{right}%
\renewcommand{\gbcap}{base's boxes}%
\renewcommand{\gbb}{0.126/0.504/0.152/0.717, 0.174/0.504/0.200/0.704,
  0.216/0.504/0.242/0.683, 0.258/0.504/0.283/0.666, 0.299/0.504/0.325/0.653,
  0.341/0.504/0.366/0.640, 0.382/0.504/0.405/0.623, 0.424/0.504/0.446/0.610,
  0.462/0.504/0.485/0.598, 0.498/0.504/0.520/0.572}%
\renewcommand{\gbo}{0.539/0.324/0.571/0.653, 0.498/0.303/0.530/0.615}%
\trrow{b}{\trnext{a}}{tr_candles}{0.6875/0.562/0.999/0.796}
  {Are the blue candles on the right or on the left?}{right}
  {``There are blue candles. Let's find them. One blue candle is at
    [727, 680, 735, 730]. Another blue candle is at [742, 680, 750, 727]. \ldots\
    Leftmost candles: Blue. Middle candles: Blue. Rightmost candles: Pink.''}{left}
  {``Most candles are white or yellowish. There are a couple of blue ones. One is
    near the right edge of the candle cluster [856, 638, 866, 715]. Another is
    slightly to the left of that one, but still on the right side [843, 633, 853,
    706].''}

\trsep{b}
\renewcommand{\trbench}{V$^*$}\renewcommand{\trans}{(B) right}%
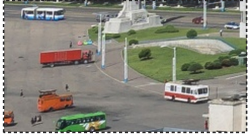
\renewcommand{\gbb}{0.043/0.035/0.241/0.150, 0.122/0.335/0.360/0.473}%
\renewcommand{\gbo}{0.122/0.381/0.360/0.554, 0.657/0.612/0.836/0.762}%
\trrow{c}{\trnext{b}}{tr_vstar}{0.008/0.425/0.078/0.585}
  {Is the white truck on the left or right side of the red truck? \quad
   (A) left \ (B) right}{(B) right}
  {``There is a white truck [19, 463, 29, 473]. There is a red truck
    [23, 489, 35, 501]. They are very close. \ldots\ So the white truck is to the
    left (smaller x).''}{(A) left}
  {``I see a red truck at [23, 493, 35, 508]. I see a white truck at
    [50, 513, 59, 526]. \ldots\ The white truck is at a higher x-value (further
    right) than the red truck.''}

\node[fit=(abg)(bbg)(cbg), inner sep=0pt] (all) {};
\pgfresetboundingbox
\useasboundingbox (all.north west) rectangle (all.south east);

\end{tikzpicture}
\caption{\textbf{Thinking traces and stated locations on GQA and V$^*$.}
For three questions, we show the original image, ground truth answer, and the Qwen3.5-4B base model's and \ours traces,
both with thinking enabled. The zoomed regions and overlaid boxes
are added for visualization; the boxes plot coordinates stated in the
traces and were not shown to the models. Our model's stated locations
align with the objects in the questions. The Qwen3.5-4B places its white-truck
box on the wrong vehicle in (c), puts ten evenly spaced boxes along one line,
none of them on a candle, in (b), and states no bounding box at all in (a).}
\label{fig:traces_ground}

\end{figure}

\paragraph{Thinking traces.}
The three CVBench examples in \autoref{fig:traces_cvbench} show how the models interpret distance in a scene. In example a), the base model treats the alignment of the lamp and door in the image as evidence that they are close, while our model places the door on the far wall and the table in the foreground. The example b) and c) show a similar difference in how the models use the scene layout to compare objects. In \autoref{fig:traces_ground}, the stated locations make the contrast more explicit: in example a) \ours identifies the goat beside the walking person, in example b) \ours places the blue candles on the right, and in example c) locates the white truck relative to the red one. The base model instead calls the goat a dog, gives candle wrong locations, and identifies the wrong vehicle.

\begin{figure}[p]
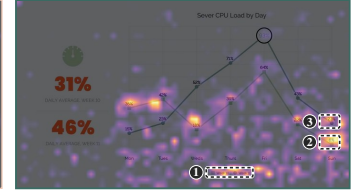

\centering
\IfFileExists{ve_common.tex}{\input{ve_common}}{\input{figures/ve_common}}%
\graphicspath{{images/}{figures/images/}}%
\renewcommand{\vencols}{3}%

\begin{tikzpicture}[ve]

\velarge{a}{(0,0)}
  {What is the label of the third highest line in year 2018?}
  {United States}{evochart_000047}
  {base/Europe/0/{attends \rin{1} \rout{2} \rout{3}}, visionopd/Europe/0/{attends \rin{1} \rout{2} \rout{3}}, ours/United States/1/{attends \rin{1} \rin{2} \rin{3}}}
\velargeheads{a}{Qwen3.5-4B (L27), Vision-OPD (L15), \ours (L15)}
\veregions{ain}{0.03/0.10/0.26/0.145/1, 0.83/0.19/0.89/0.27/2, 0.83/0.64/0.89/0.76/3}
\veregions{ap1}{0.03/0.10/0.26/0.145/1, 0.83/0.19/0.89/0.27/2, 0.83/0.64/0.89/0.76/3}
\veregions{ap2}{0.03/0.10/0.26/0.145/1, 0.83/0.19/0.89/0.27/2, 0.83/0.64/0.89/0.76/3}
\veregions{ap3}{0.03/0.10/0.26/0.145/1, 0.83/0.19/0.89/0.27/2, 0.83/0.64/0.89/0.76/3}

\vesep{a}
\velarge{b}{\venext{a}}
  {What is the value of the blue line when the green line reaches the lowest value?}
  {23}{evochart_000087}
  {base/18/0/{attends \rout{1} \rin{2} \rin{3}}, visionopd/18/0/{attends \rout{1} \rout{2} \rin{3}}, ours/23/1/{attends \rin{1} \rin{2} \rin{3}}}
\velargeheads{b}{Qwen3.5-4B (L27), Vision-OPD (L15), \ours (L15)}
\veregions{bin}{0.57/0.89/0.71/0.94/1, 0.91/0.72/0.97/0.79/2, 0.91/0.61/0.97/0.68/3}
\veregions{bp1}{0.57/0.89/0.71/0.94/1, 0.91/0.72/0.97/0.79/2, 0.91/0.61/0.97/0.68/3}
\veregions{bp2}{0.57/0.89/0.71/0.94/1, 0.91/0.72/0.97/0.79/2, 0.91/0.61/0.97/0.68/3}
\veregions{bp3}{0.57/0.89/0.71/0.94/1, 0.91/0.72/0.97/0.79/2, 0.91/0.61/0.97/0.68/3}

\node[fit=(abg)(bbg), inner sep=0pt] (all) {};
\pgfresetboundingbox
\useasboundingbox (all.north west) rectangle (all.south east);

\end{tikzpicture}
\caption{\textbf{Enlarged EvoChart examples.}
The EvoChart example from \autoref{fig:visual_evidence}, enlarged as well as an additional EvoChart example. We show each visual input, question, ground-truth answer, and attention maps from the response tokens to the image for Qwen3.5-4B, Vision-OPD, and \ours. Numbered
dashed boxes mark the regions each question’s answer depends on, drawn identically on every map. Under each map, a region is ticked when it receives at least twice its area share
of the attention.}
\label{fig:visual_evidence_evochart_large}
\end{figure}

\begin{figure}[p]
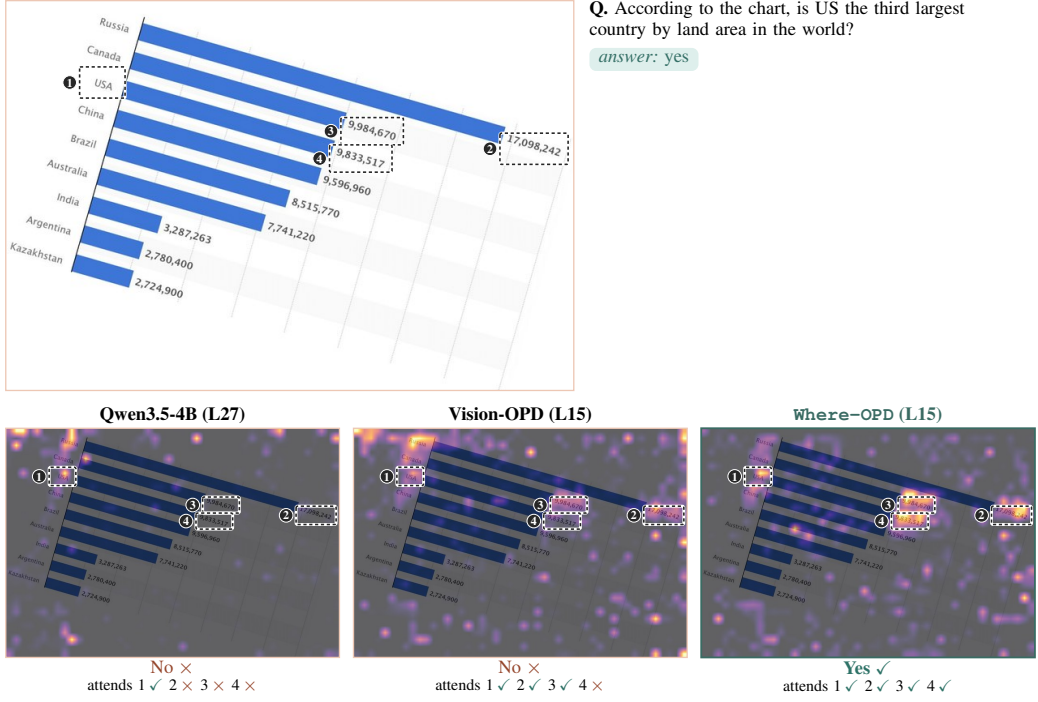

\centering
\IfFileExists{ve_common.tex}{\input{ve_common}}{\input{figures/ve_common}}%
\graphicspath{{images/}{figures/images/}}%
\renewcommand{\vencols}{3}%

\begin{tikzpicture}[ve]

\velarge{a}{(0,0)}
  {According to the chart, is US the third largest country by land area in the world?}
  {yes}{hallusion_000033}
  {base/No/0/{attends \rin{1} \rout{2} \rout{3} \rout{4}}, visionopd/No/0/{attends \rin{1} \rin{2} \rin{3} \rout{4}}, ours/Yes/1/{attends \rin{1} \rin{2} \rin{3} \rin{4}}}
\velargeheads{a}{Qwen3.5-4B (L27), Vision-OPD (L15), \ours (L15)}
\veregions{ain}{0.13/0.17/0.21/0.25/1, 0.87/0.34/0.99/0.42/2, 0.59/0.30/0.70/0.37/3, 0.57/0.37/0.68/0.44/4}
\veregions{ap1}{0.13/0.17/0.21/0.25/1, 0.87/0.34/0.99/0.42/2, 0.59/0.30/0.70/0.37/3, 0.57/0.37/0.68/0.44/4}
\veregions{ap2}{0.13/0.17/0.21/0.25/1, 0.87/0.34/0.99/0.42/2, 0.59/0.30/0.70/0.37/3, 0.57/0.37/0.68/0.44/4}
\veregions{ap3}{0.13/0.17/0.21/0.25/1, 0.87/0.34/0.99/0.42/2, 0.59/0.30/0.70/0.37/3, 0.57/0.37/0.68/0.44/4}

\node[fit=(abg), inner sep=0pt] (all) {};
\pgfresetboundingbox
\useasboundingbox (all.north west) rectangle (all.south east);

\end{tikzpicture}
\caption{\textbf{Enlarged HallusionBench example.}
The HallusionBench example from \autoref{fig:visual_evidence}, enlarged.  We show each visual input, question, ground-truth answer, and attention maps from the response tokens to the image for Qwen3.5-4B, Vision-OPD, and \ours. Numbered
dashed boxes mark the regions each question’s answer depends on, drawn identically on every map. Under each map, a region is ticked when it receives at least twice its area share
of the attention.}
\label{fig:visual_evidence_hallusion_large}
\end{figure}

\begin{figure}[p]
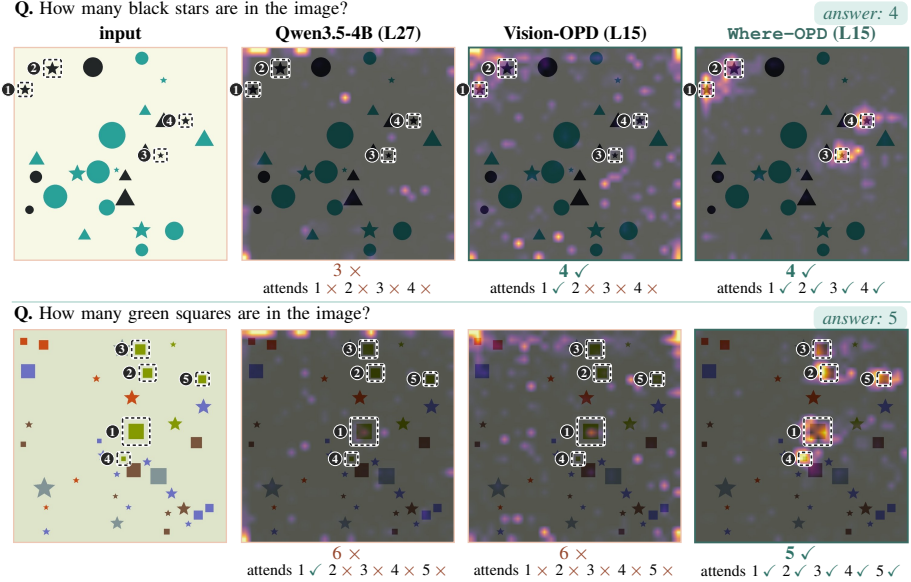

\centering
\IfFileExists{ve_common.tex}{\input{ve_common}}{\input{figures/ve_common}}%
\graphicspath{{images/}{figures/images/}}%
\renewcommand{\inw}{28mm}\renewcommand{\panw}{28mm}%
\renewcommand{\vencols}{3}%

\newcommand{\targetsA}{0.830/0.341/0.044, 0.942/0.774/0.047, 0.045/0.808/0.015}
\newcommand{\targetsB}{0.576/0.479/0.051, 0.630/0.203/0.031, 0.596/0.092/0.033,
                       0.517/0.607/0.015, 0.888/0.232/0.025}

\begin{tikzpicture}[ve]

\veblock[3.8mm]{a}{(0,0)}
  {How many black stars are in the image?}{4}{counting_000139}
  {base/3/0/{attends \rout{1} \rout{2} \rout{3} \rout{4}},
   visionopd/4/1/{attends \rin{1} \rout{2} \rout{3} \rout{4}},
   ours/4/1/{attends \rin{1} \rin{2} \rin{3} \rin{4}}}
\veheads{a}{Qwen3.5-4B (L27), Vision-OPD (L15), \ours (L15)}
\foreach \pnl in {ain,ap1,ap2,ap3}{\veregions{\pnl}{0.019/0.165/0.084/0.230/1, 0.139/0.057/0.221/0.139/2, 0.658/0.477/0.718/0.537/3, 0.776/0.314/0.836/0.374/4}}

\vesep{a}
\veblock{b}{\venext{a}}
  {How many green squares are in the image?}{5}{counting_000195}
  {base/6/0/{attends \rin{1} \rout{2} \rout{3} \rout{4} \rout{5}},
   visionopd/6/0/{attends \rout{1} \rout{2} \rout{3} \rout{4} \rout{5}},
   ours/5/1/{attends \rin{1} \rin{2} \rin{3} \rin{4} \rin{5}}}
\foreach \pnl in {bin,bp1,bp2,bp3}{\veregions{\pnl}{0.512/0.415/0.640/0.543/1, 0.591/0.164/0.669/0.242/2, 0.555/0.051/0.637/0.133/3, 0.487/0.577/0.547/0.637/4, 0.857/0.201/0.919/0.263/5}}

\pgfresetboundingbox
\useasboundingbox (abg.north west) rectangle (bbg.south east);

\end{tikzpicture}
\caption{\textbf{Attention on target objects in synthetic counting scenes.}
Two held-out examples with visual input, question, ground-truth answer, and attention maps from the response tokens to the image for Qwen3.5-4B, Vision-OPD, and \ours.
Numbered
dashed boxes mark the regions each question’s answer depends on, drawn identically on every map. Under each map, a region is ticked when it receives at least twice its area share
of the attention. The
displayed layers are selected as described in \autoref{tab:layer_selection}.}
\label{fig:visual_evidence_counting}
\end{figure}

\paragraph{Visual evidence.}
The enlarged EvoChart examples in \autoref{fig:visual_evidence_evochart_large} show two questions that require combining information from different parts of a chart: ranking the lines at a given year (a), and finding the green line's minimum before reading the blue line (b). In both cases, \ours answers correctly and its attention covers all regions of importance marked in the figure for visualization purposes, while the base model and Vision-OPD omit some of them. The enlarged HallusionBench example (\autoref{fig:visual_evidence_hallusion_large}) shows a similar pattern: our model attends to the USA label and the three land-area values needed to check its rank, and correctly answers the question.

\subsection{Layer selection for attention map visualization}
\label{sec:exp:attention}

Let $a$ be the attention from the response tokens to the $H{\times}W$ visual tokens of one image, averaged over heads and over the response tokens and renormalized over the visual tokens alone. For a token set $\mathcal{S}$, enrichment $\mathrm{enr}(\mathcal{S})$
compares the attention assigned to $\mathcal{S}$ with the fraction of
visual tokens it contains:
\begin{equation}
    \mathrm{enr}(\mathcal{S}) = \frac{\sum_{(i,j) \in \mathcal{S}} a_{ij}}{|\mathcal{S}| / HW}.
    \label{eq:enrichment}
\end{equation}
An enrichment of $1.0$ means the targets receive exactly their area share, and higher values mean attention concentrated on the evidence. 

We measure attention enrichment on target objects (objects referred to by the question) at each of the full-attention layers of Qwen3.5-4B using a separate selection split of 158 held-out synthetic scenes (\autoref{tab:layer_selection}). The base model's enrichment peaks at L27. Vision-OPD and our model both peak at L15, where our model has higher enrichment. We use these selected layers for attention visualizations.

\paragraph{Attention to target objects.} We visualize two held-out counting examples (\autoref{fig:visual_evidence_counting}), our model attends to all marked target objects and gives the correct count, while the base model and Vision-OPD attend to fewer targets.

\end{document}